\documentclass[10pt,twocolumn,letterpaper]{article}

\usepackage{cvpr}              

\usepackage[dvipsnames]{xcolor}

\usepackage{times}
\usepackage{epsfig}
\usepackage{graphicx}
\usepackage{amsmath}
\usepackage{amssymb}
\usepackage{booktabs}
\usepackage[ruled,vlined]{algorithm2e}
\usepackage{comment}
\usepackage{multirow}
\usepackage{enumitem}
\usepackage{microtype}
\usepackage{listings}

\usepackage[utf8]{inputenc} 
\usepackage[T1]{fontenc}    
\usepackage{xcolor, color, colortbl}         
\definecolor{Highlight}{HTML}{39b54a}  

\usepackage[symbol]{footmisc}

\definecolor{cvprblue}{rgb}{0.21,0.49,0.74}
\usepackage[pagebackref,breaklinks,colorlinks,citecolor=cvprblue]{hyperref}

\usepackage[capitalize]{cleveref}
\crefname{section}{Sec.}{Secs.}
\Crefname{section}{Section}{Sections}
\Crefname{table}{Table}{Tables}
\crefname{table}{Tab.}{Tabs.}

\def\paperID{162} 
\def\confName{3DV\xspace}
\def\confYear{2024\xspace}

\title{Test-Time Augmentation for 3D Point Cloud Classification and Segmentation}

\author{Tuan-Anh Vu\textsuperscript{1} \quad 
Srinjay Sarkar\textsuperscript{2}\footnote[1]{co-first author} \quad  
Zhiyuan Zhang\textsuperscript{3}\footnote[2]{corresponding author} \quad
Binh-Son Hua\textsuperscript{4} \quad 
Sai-Kit Yeung\textsuperscript{1} \\ \\
\textsuperscript{1}The Hong Kong University of Science and Technology \quad \textsuperscript{2}VinAI Research \\  \textsuperscript{3}Singapore Management University \quad \textsuperscript{4}Trinity College Dublin 
}

\begin{document}
\maketitle
\footnotetext[1]{co-first author.}
\footnotetext[2]{corresponding author.}
\begin{abstract}
Data augmentation is a powerful technique to enhance the performance of a deep learning task but has received less attention in 3D deep learning. It is well known that when 3D shapes are sparsely represented with low point density, the performance of the downstream tasks drops significantly. This work explores test-time augmentation (TTA) for 3D point clouds. We are inspired by the recent revolution of learning implicit representation and point cloud upsampling, which can produce high-quality 3D surface reconstruction and proximity-to-surface, respectively. Our idea is to leverage the implicit field reconstruction or point cloud upsampling techniques as a systematic way to augment point cloud data. Mainly, we test both strategies by sampling points from the reconstructed results and using the sampled point cloud as test-time augmented data. We show that both strategies are effective in improving accuracy. We observed that point cloud upsampling for test-time augmentation can lead to more significant performance improvement on downstream tasks such as object classification and segmentation on the ModelNet40, ShapeNet, ScanObjectNN, and SemanticKITTI datasets, especially for sparse point clouds.
\end{abstract}    
\section{Introduction}
\label{sec:intro}

Point-based representation is of great importance to computer graphics and computer vision. In the modern era of deep learning, neural networks can be designed to learn features from point clouds, facilitating 3D perception tasks such as object classification, object detection, and semantic segmentation in many downstream applications. Nevertheless, such evolutions still leave 3D perception a challenging and unsolved problem. A typical disadvantage of point-based representation is that surface information is implied by point density and orientation, if any. Due to such ambiguity, techniques for data augmentation on point clouds are relatively scarce and challenging to design. 

Recent advances in using neural networks to represent 3D data have opened new opportunities to revise and explore this problem from a new perspective~\cite{mescheder2019occupancy,peng2020convolutional,sapcu}. One type of method is the so-called neural implicit representation based on the idea of training a neural network that can return queries of the 3D space from input coordinates~\cite{park2019deepsdf,mescheder2019occupancy,sitzmann2019scene,peng2020convolutional}. Particularly, one can train a neural network to encode a 3D point to various attributes such as occupancy, color, or a general feature vector. The power of a neural implicit representation is that the queries can be performed at arbitrary points, and no special mechanism is required for value interpolation. Another type of methods~\cite{yu2018pu,li2021point, sapcu} employs upsampling to achieve both distribution uniformity and proximity-to-surface. The advantages of the upsampling-based method lie in self-supervision and more uniformly distributed dense representation without the need of surface ground truth.

In this work, we investigate both types of strategies and leverage them as a systematic way for data augmentation at test time. Particularly, for implicit representation, we leverage the convolutional occupancy network \cite{peng2020convolutional} to encode the 3D point clouds to a regular grid representation that allows the interpolation of features at an arbitrary location. For the upsampling-based method, we employ the self-upsampling method~\cite{sapcu} to obtain a dense and uniformly distributed proximity-to-surface point cloud. We propose an effective technique to aggregate features of the original and augmented point clouds to generate the final prediction. We select the task of object classification and semantic segmentation as the downstream task to validate our augmentation technique, as they play a key role in many practical applications, including perception in robotics and autonomous driving. We experiment with point cloud data from ModelNet40~\cite{wu20153d}, ShapeNet~\cite{chang2015shapenet}, ScanObjectNN~\cite{uy-scanobjectnn-iccv19} and SemanticKITTI~\cite{behley2019semantickitti} dataset, which demonstrates significant performance improvement. 

In summary, our key contributions are:
\begin{itemize}[leftmargin=*]
    \item We analyze and compare existing reconstruction approaches, including surface-based sampling and point cloud upsampling for test-time augmentation.
    \item We propose a test-time augmentation method for 3D point cloud deep learning, which is suitable for both approaches; 
    \item We identified a self-supervised point cloud upsampling method as a robust method for our test-time augmentation. It uses the proximity-to-surface cues to sample augmented point clouds.
    \item Extensive experiments and analysis prove the effectiveness of our augmentation method on two downstream tasks, including object classification and semantic segmentation on synthetic and real-world datasets.
\end{itemize}

\section{Related Works}
\label{sec:related}

\paragraph{3D Deep Learning.}
Early methods usually convert the irregular and sparse 3D points into multiple regular 2D views~\cite{xie2015projective, kalogerakis20173d, dai20183dmv} or 3D voxels~\cite{wu20153d, wang2019voxsegnet,huang2016point,dai2017scannet,tchapmi2017segcloud}. Despite the improvements gained by performing CNN on these regular structures, these methods usually suffer information loss and high computational costs.

Point cloud is a universal representation for 3D data. PointNet~\cite{qi2017pointnet} is the pioneering work that can process 3D points directly by symmetric functions and max-pooling to extract global features. To capture local features, PointNet++~\cite{qi2017pointnet++} performs hierarchical PointNets on different scales. In recent years, various convolution operators and networks have been proposed for point clouds, such as PointCNN~\cite{li2018pointcnn}, SpiderCNN~\cite{xu2018spidercnn}, DGCNN~\cite{wang2018edgeconv}, and ShellNet~\cite{zhang-shellnet-iccv19} with the supreme performance achieved on classification, retrieval, and segmentation tasks. There are also approaches~\cite{huang2018recurrent,wang2018deep,graham20183d,hu2020randla,xu2020squeezesegv3} specially designed for semantic segmentation. Huang et al.~\cite{huang2018recurrent} learn the local structure, particularly for semantic segmentation, by applying learning algorithms from recurrent neural networks. SPG~\cite{landrieu2018large} constructs graphs between coarsely segmented super-points for large-scale point cloud semantic segmentation. 

To balance the efficiency and accuracy, hybrid works~\cite{le2018pointgrid,liu2019point,zhang2020deep} utilize the characteristics of multiple representations. PointGrid~\cite{le2018pointgrid} assigns fix number of points in each grid cell, making the conventional CNN feasible. While it runs fast, the accuracy is still not high. PVCNN~\cite{liu2019point} represents the input in points and performs the convolutions in voxels with a superior performance achieved than sole point or voxel representations. To handle large-scale lidar point cloud data,  FusionNet~\cite{zhang2020deep} divides the input into voxels and extracts features from both voxels and inner points. 

\noindent\textbf{Neural 3D Reconstruction.}
3D reconstruction works can be classified into four categories in terms of the output representation: voxel-based, point-based, mesh-based, and implicit function-based methods.

Similar to semantic segmentation, voxel is also a popular representation for 3D reconstruction~\cite{choy20163d, wu2016learning, wu20153d}. In the category, voxel grids are used to store either occupancy that encodes whether the voxel is occupied or not~\cite{wu20153d, choy20163d} or SDF information that holds signed projective distances from voxels to the closest surfaces~\cite{dai2017shape, liao2018deep, stutz2018learning}. However, as mentioned in the segmentation works, such methods inherit the limitations of high memory costs.

Another line of works output point clouds directly for 3D reconstruction~\cite{fan2017point, lin2018learning, prokudin2019efficient, yang2019pointflow}. These methods design generative models to produce dense points for scene representation. Despite the efficiency, the generated points cannot sufficiently represent complicated surfaces as there is no topology between the points.

Mesh is another popular output representation for 3D reconstruction. In this category, some works deform shapes with simple topology to more complicated shapes, which usually constrain to certain fixed templates~\cite{kanazawa2018end, ranjan2018generating} or topologies~\cite{sinha2016deep, ben2018multi}. To reconstruct a shape of arbitrary topology, AtlasNet~\cite{groueix2018atlasnet} warps multiple 2D planes into 3D shapes. Despite the superior results, this method can result in self-intersecting mesh faces.

To overcome the limitations of the above explicit representations (voxel, point, mesh), more recent works focus on implicit representations that employ occupancy~\cite{mescheder2019occupancy, peng2020convolutional} and distance field~\cite{park2019deepsdf, chabra2020deep} with a neural network to infer an occupancy probability or distance value for the input 3D points. As implicit representation models shape continuously, more detail is preserved, and more complicated shape topologies can be obtained. In this work, we also employ implicit representation to aim to augment a point cloud for various downstream tasks.

\begin{figure*}[t]
    \centering
    \includegraphics[width=\linewidth]{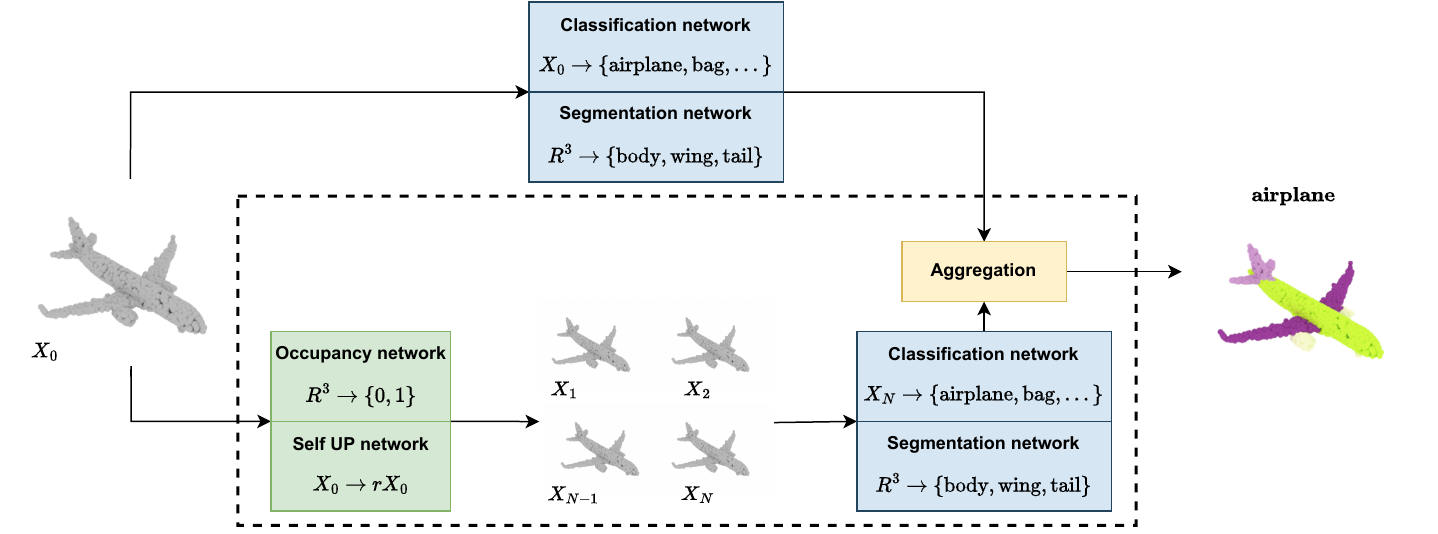}
    \caption{Illustration of our test-time augmentation method for point clouds downstream tasks such as classification and segmentation. We view the input point cloud as a noisy estimate of a latent surface and propose using an implicit field represented by an occupancy network or a point cloud upsampling network to sample augmented point clouds so that the point clouds share the same underlying surfaces. We then perform the downstream task on each point cloud and aggregate the point features to produce the final result.}
    \label{fig:our_tta}
\end{figure*}

\noindent\textbf{Point Cloud Upsampling.}
Point cloud upsampling can produce a dense, uniform, and complete point cloud from a sparse and noisy complete with or without missing parts. 

\textit{Traditional Point Cloud Upsampling:} A seminal point cloud upsampling algorithm is to interpolate points as vertices of a Voronoi diagram~\cite{Alexa2003ComputingAR}. \cite{lipman_lop} later proposed an algorithm by introducing the locally optimal linear projector for surface reconstruction and using it to project a set of points onto the input point cloud. This work was followed by \cite{huang_wlop}, who proposed a weighted locally linear operator in order to make the point cloud distribution more even. \cite{huang_edgeaware} introduces an edge-aware resampling method by sampling points on edge and calculating the normals at those points. All of the above-mentioned methods are not data-driven and thus heavily rely on priors like normal estimation.

\textit{Deep-Learning Based Point Cloud Upsampling:}
PU-Net~\cite{yu2018pu} was the first deep learning-based point cloud upsampling method that used a multi-branch feature expansion module to extract multi-scale features and expand a point cloud in the feature space. This was followed by EC-Net~\cite{ec-net}, which achieves edge-aware point cloud upsampling by learning distance features obtained by the perturbation of the generated point cloud relative to the input point cloud. In this work, we propose to use point cloud upsampling as a test-time augmentation technique.

\vspace{0.05in}
\noindent\textbf{Data Augmentation and Test-Time Augmentation.} 
In modern deep learning, large-scale data is often required for training deep neural networks; however, acquiring a large amount of data is a thorough and prohibitively expensive process.
Data augmentation is a common but useful technique to scale up the data artificially. In image classification, popular data augmentation includes simple transformations of the images, including rotations, flipping, cropping, etc.~\cite{krizhevsky2012imagenet,howard2013improvements,szegedy2015going,simonyan2015very,he2016residual,howard2013improvements}. In 3D deep learning, traditional methods (e.g., PointNet~\cite{qi2017pointnet} and PointNet++~\cite{qi2017pointnet++}) utilize similarity transformations such as random rotations, scaling, and jittering for data augmentation during training. Similarity transformations are also used to augment real-world data to build ScanObjectNN~\cite{uy-scanobjectnn-iccv19}, a real-world dataset for object classification. Research efforts for more sophisticated augmentation techniques for 3D point clouds are relatively scarce. Recently, PointMixup~\cite{chen2020pointmixup} is proposed to mix two point clouds based on shortest path linear interpolation; PointAugment~\cite{li2020pointaugment} uses adversarial learning to seek augmented point clouds satisfying a given classifier. PPBA~\cite{cheng2020population} automates the design of augmentation policies for the specific task of 3D object detection. Apart from train-time data augmentation, in this work, we assume that pre-trained models for specific downstream tasks are already given and investigate test-time augmentation techniques~\cite{ayhan2018} that can boost overall performance. 

Particularly, Test-time augmentation (TTA) first transforms the input, then perform predictions on the augmented versions of the input, and finally combines the prediction results. This strategy is common to image classification with simple augmentation policies such as flipping, cropping, and scaling~\cite{howard2013improvements}. More sophisticated methods involve learning and optimizing augmentation policies~\cite{sato2015apac,kim2020learning,lyzhov2020greedy,shanmugam2020testtime}, or learning to combine predictions~\cite{shanmugam2020testtime}. Beyond images, TTA has also been applied to medical image segmentation~\cite{moshkov2020test,wang2019aleatoric} and text recognition~\cite{li2019show}. For point clouds, however, we are unaware of any recent method tailored to test-time augmentation and augmentation policies. 

\section{Our Method}
\label{sec:method}

\subsection{Overview}
Given a point set $\{\mathbf{p}_i \}_{i=1}^n$ with $\mathbf{p}_i \in \mathbb{R}^3$ represented by a matrix $\mathbf{x}_0 \in \mathbb{R}^{n \times 3}$. Without loss of generality, we assume at inference, $\mathbf{x}_0$ is passed to pre-trained network $f$ for feature extraction, and the features are passed to a network $g$ for final label prediction $f(\mathbf{x}_0)$. Our goal is to achieve performance improvement in the downstream task via test-time augmentation, where the final prediction can be defined as:
\begin{align}
    g(\phi(f(\mathbf{x}_0), f(\mathbf{x}_1), f(\mathbf{x}_2), ...))
\end{align}
where $\phi$ is an aggregation function to combine multiple features resulting from the original point set $\mathbf{x}_0$ and the augmented point sets $\mathbf{x}_1$, $\mathbf{x}_2$, etc. Note that the network $f$ and $g$ are pre-trained and left untouched in test-time augmentation; only the input is augmented.

Traditionally, a simple method for test-time augmentation is jittering, which adds Gaussian noise to perturb the point cloud $\mathbf{x}_0$ to generate an augmented point cloud $\mathbf{x}_k$: 
\begin{align}
    \mathbf{x}_k = \mathbf{x}_0 + \lambda \mathbf{z}_k
\end{align}
where $\mathbf{z}_k \sim \mathcal{N}(0, I)$ is a random noise vector from a normal distribution, and $\lambda$ is a scalar value to control the noise magnitude. This simple augmentation has been widely adopted since the seminal PointNet~\cite{qi2017pointnet}. An issue of such augmentation is that it does not consider the underlying surface or point distribution because the noise $\mathbf{z}$ is independent of $\mathbf{x}_0$, resulting in marginal performance improvement in many cases. In this work, we viewpoint set $\mathbf{x}_0$ as a noisy estimate of a latent surface representation $\mathcal{S}$, and therefore, we define point cloud augmentation as the process of sampling additional point clouds $\mathbf{x}_k$ that explain the same surface. We propose to sample augmented point clouds $\mathbf{x}_k$ ($k \geq 1$) in two ways: surface sampling and point cloud up-sampling. 
The sampled point clouds can then be leveraged for downstream tasks such as classification and segmentation. Our method is visualized in Figure \ref{fig:our_tta}.

\begin{figure}[!t]
\centering
\def\sc{0.185}

\includegraphics[width=\sc\linewidth]{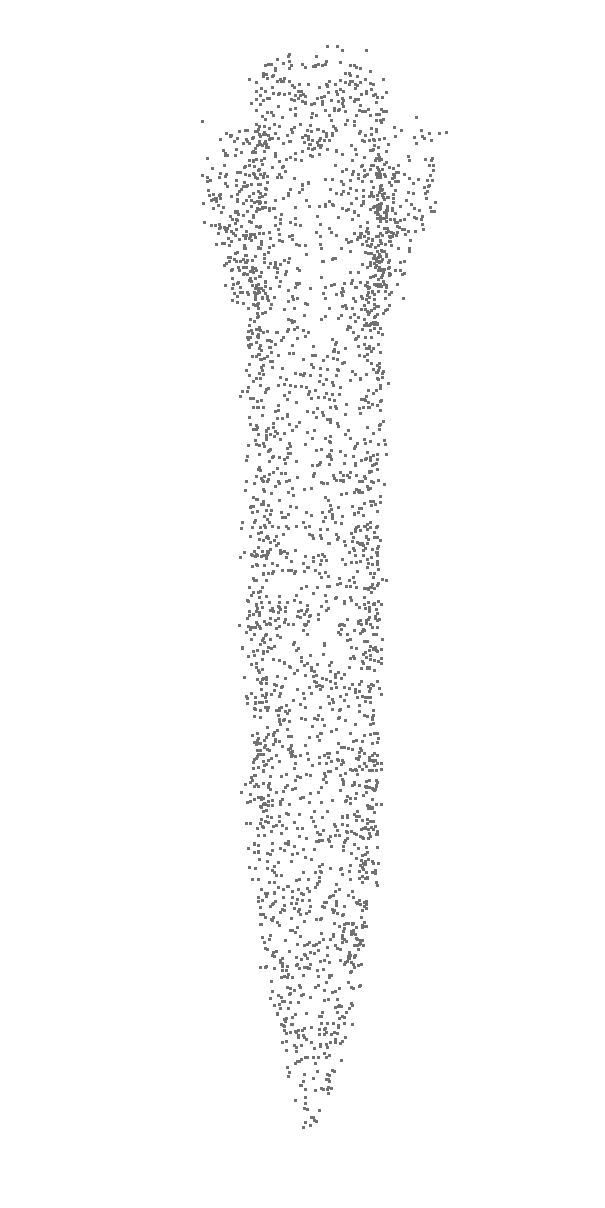}
\includegraphics[width=\sc\linewidth]{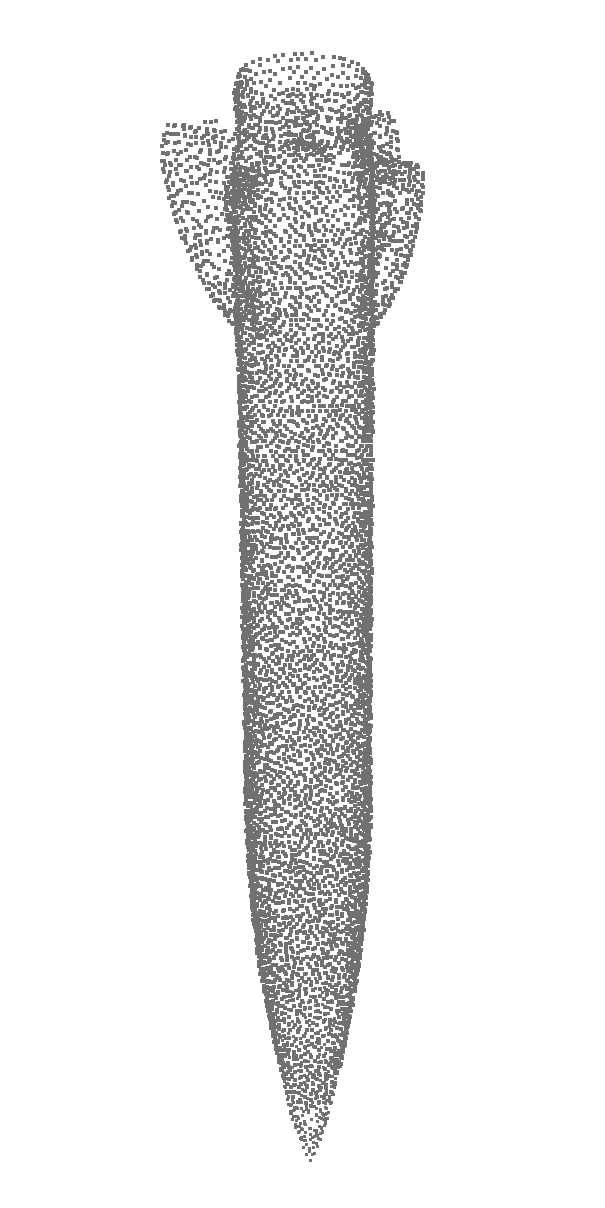}
\includegraphics[width=0.205\linewidth]{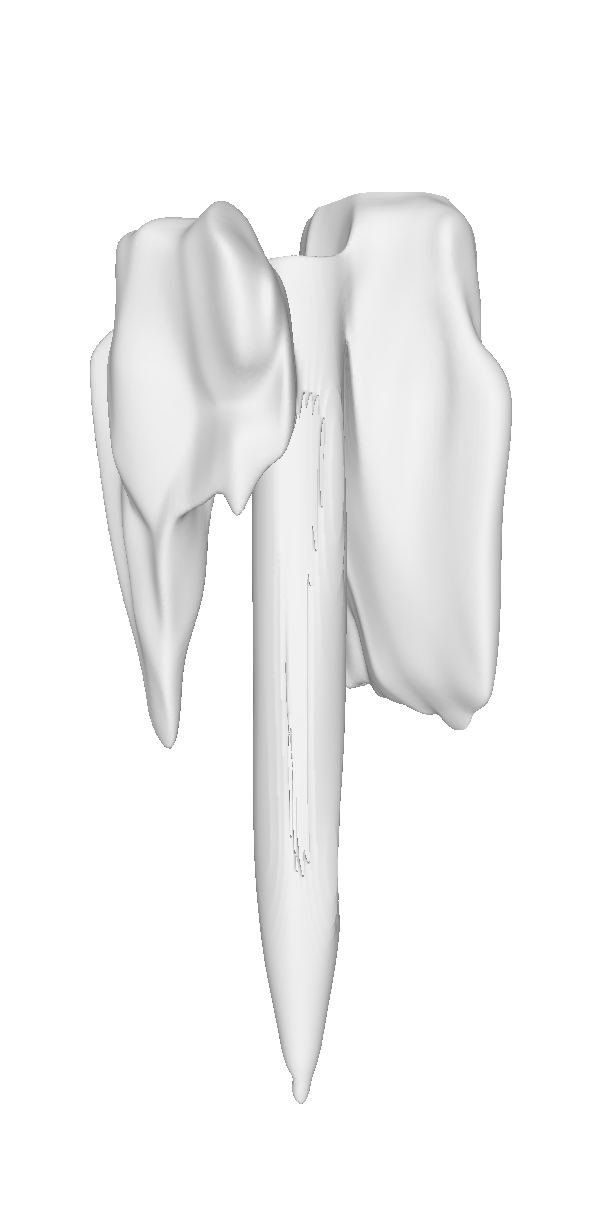}
\includegraphics[width=\sc\linewidth]{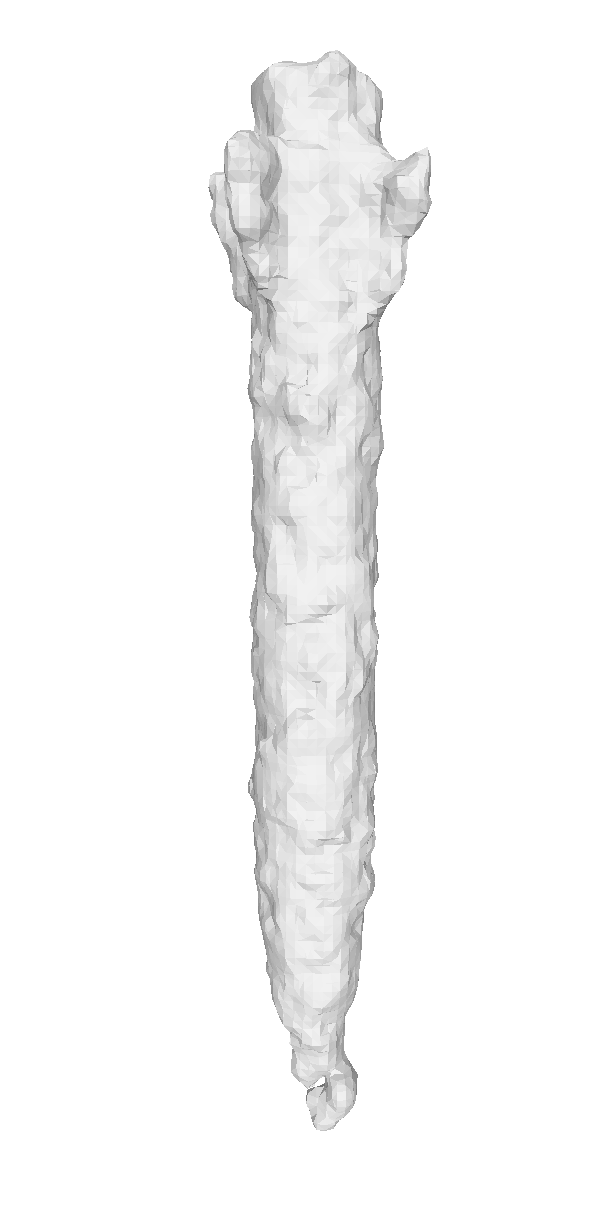}
\includegraphics[width=\sc\linewidth]{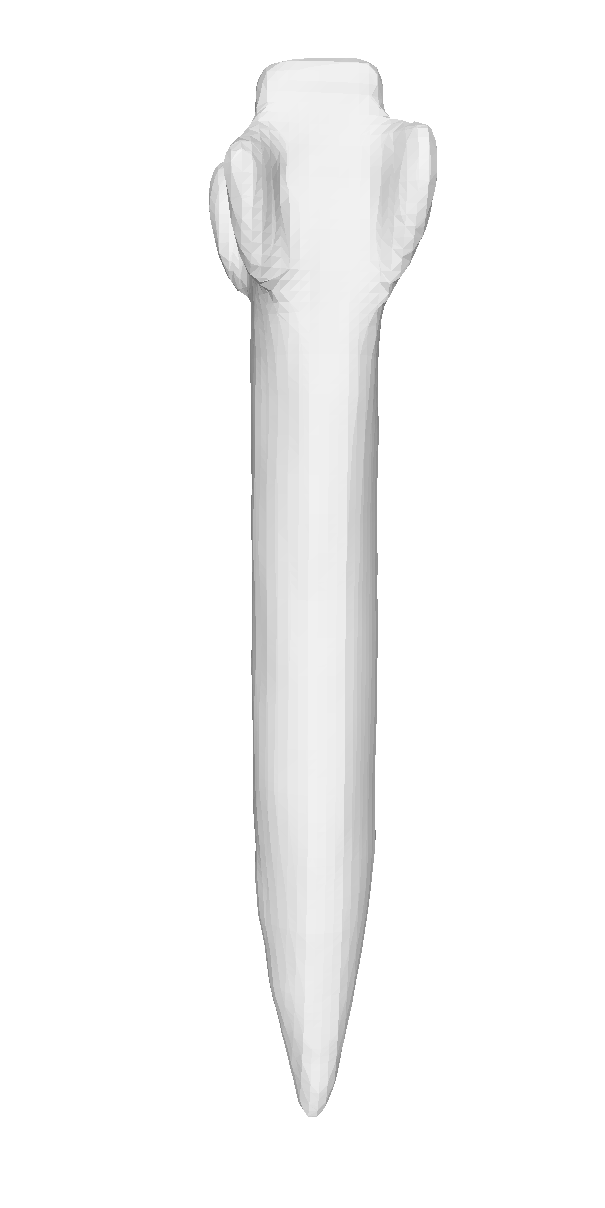}

\includegraphics[width=\sc\linewidth]{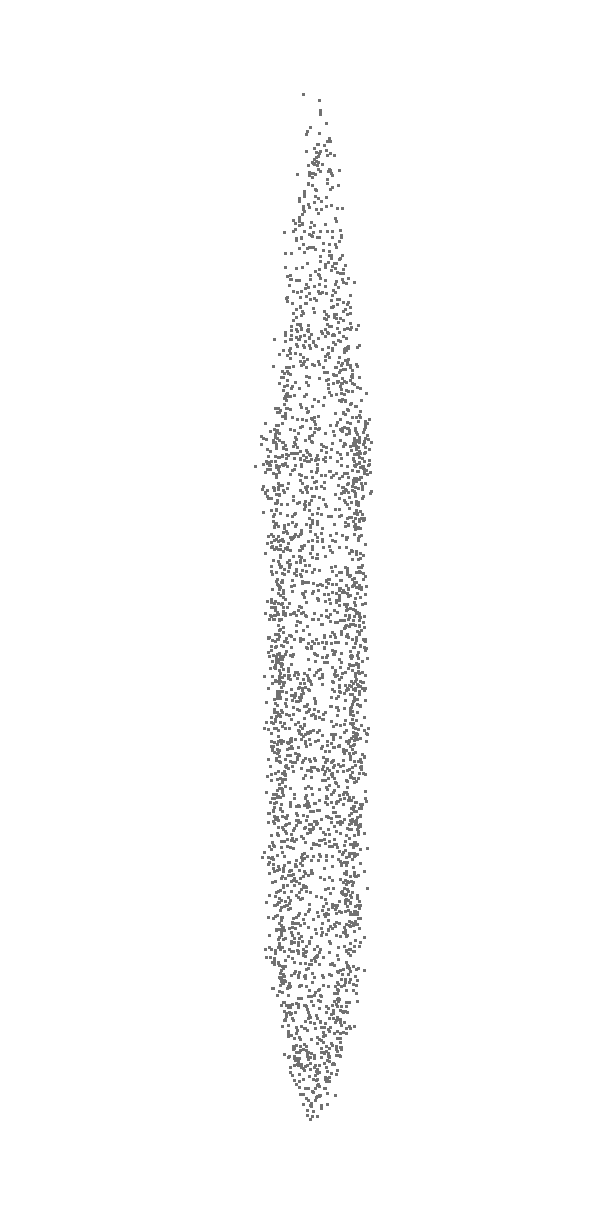}
\includegraphics[width=\sc\linewidth]{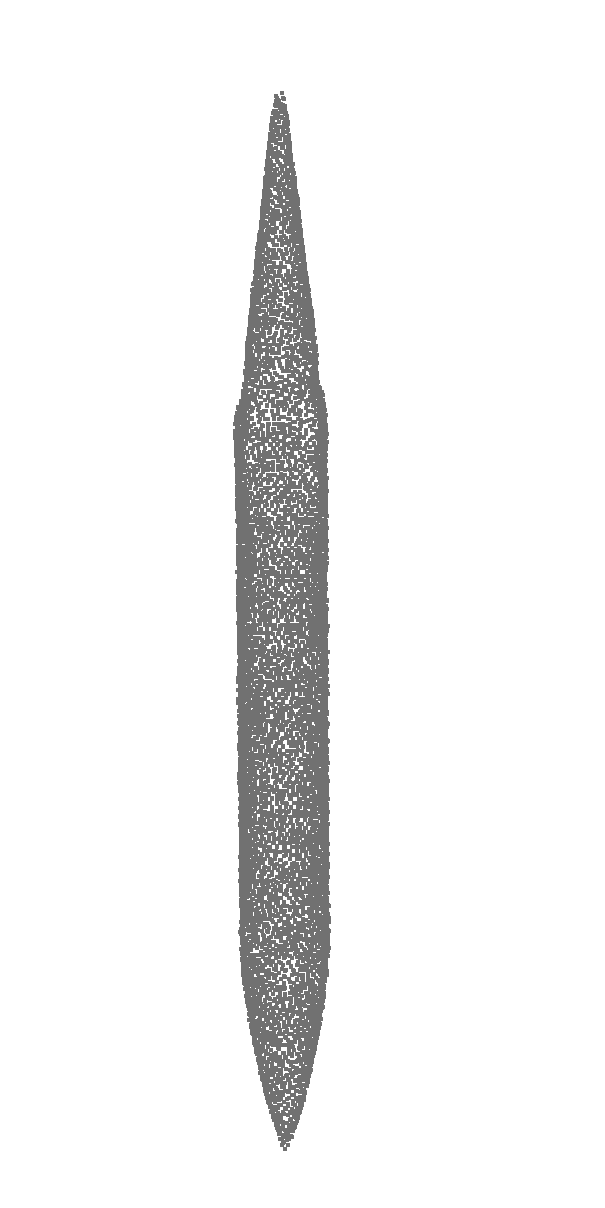}
\includegraphics[width=0.205\linewidth]{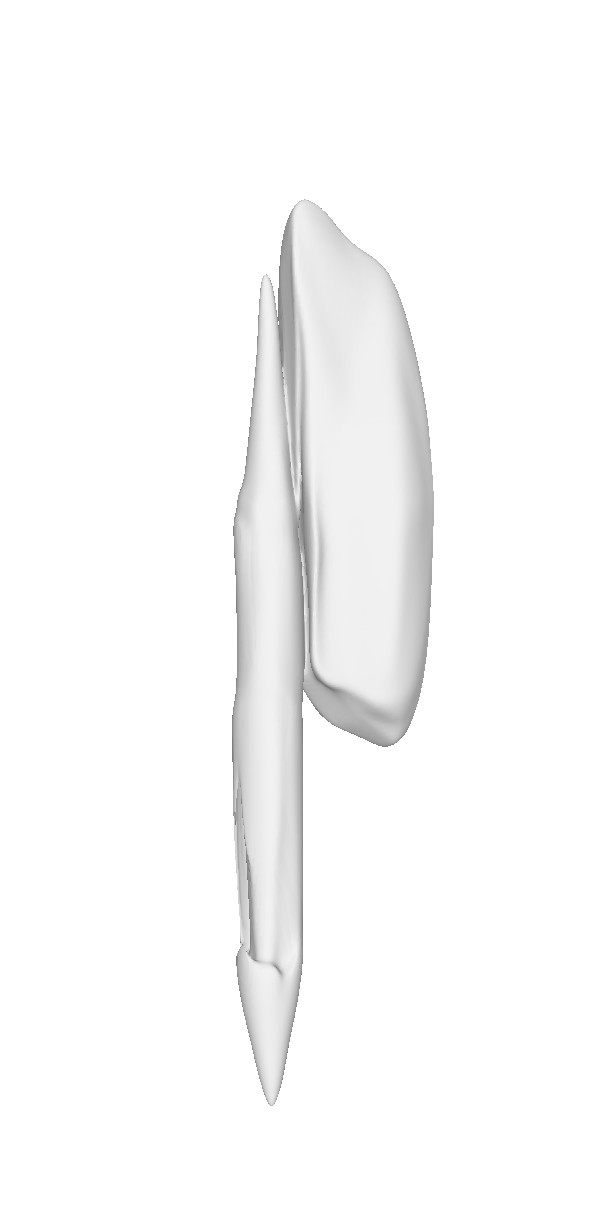}
\includegraphics[width=\sc\linewidth]{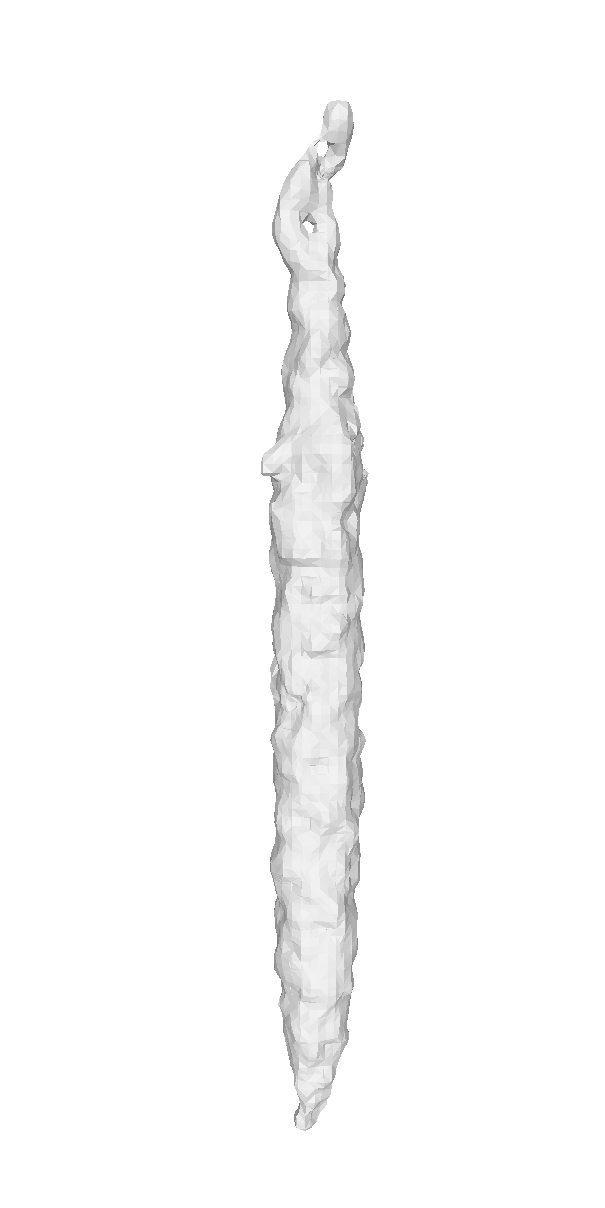}
\includegraphics[width=\sc\linewidth]{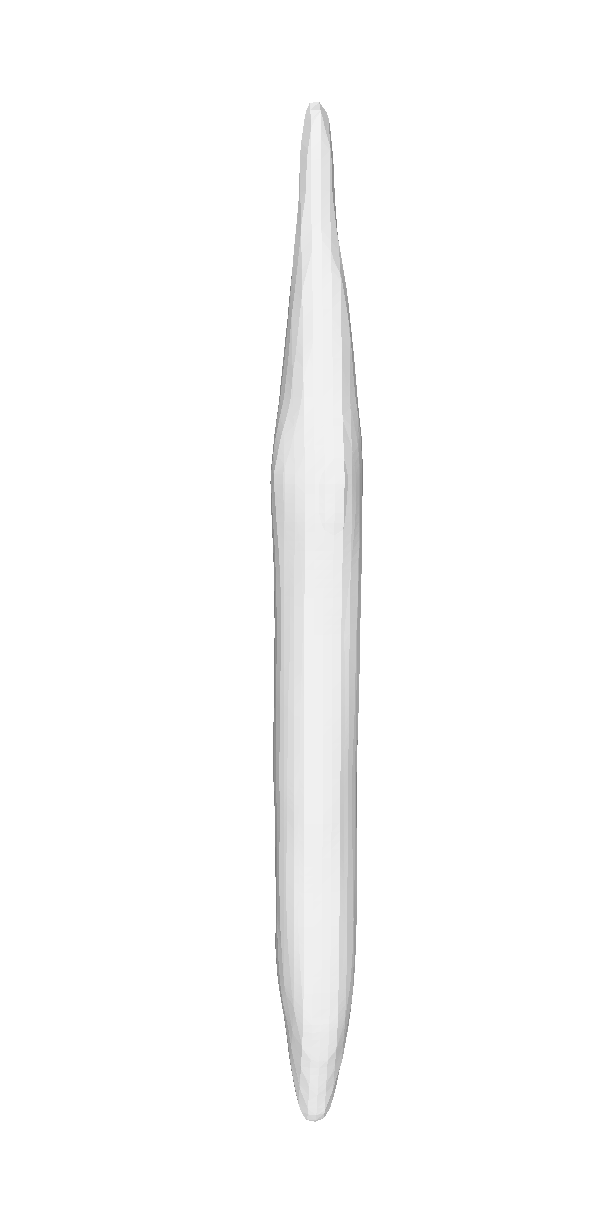}

\caption{\textbf{Visual comparison of different reconstruction and upsampling methods.} From left to right: Input, Up-sampling point clouds, Unsupervised reconstruction, Poisson reconstruction, and Supervised reconstruction. As can be seen, the shape quality of supervised reconstruction~\cite{peng2020convolutional} using neural implicit representation is finer and smoother compared to unsupervised method~\cite{Gropp2020} and Screened Poisson method~\cite{Kazhdan2006}. In addition, we can obtain a dense and uniformly distributed proximity-to-surface point cloud using Self-supervised Upsampling \cite{sapcu}, which contributes to the success of our method. Best viewed with zoom.}
\label{fig:recon_compare}
\end{figure}

In the following sections, we explain the technique for sampling augmented point clouds using an implicit representation network (Section \ref{sec:implicit_recon}) and a self-supervised point upsampling network (Section \ref{sec:self_up}). We then present downstream tasks that leverage the proposed test-time augmentation and discuss feature aggregation and final label prediction for point cloud classification and segmentation. 

\subsection{Augmentation by Implicit Field Reconstruction}
\label{sec:implicit_recon}
We are motivated by the recent advances in geometry reconstruction using neural implicit representation. The basic idea is to learn a mapping $f_\theta: \mathbb{R}^3 \longrightarrow \{0, 1\}$ using a neural network parameterized by $\theta$. This function implicitly encodes the geometry in the 3D space to allow the query of the occupancy at any point in the 3D space. To obtain the geometry explicitly, the Marching Cubes algorithm~\cite{lorensen1987marchingcubes} can generate a triangle mesh containing surfaces at zero crossings in the implicit field. Our neural implicit field is built upon the convolutional occupancy network~\cite{mescheder2019occupancy,peng2020convolutional}. The convolutional occupancy network uses a combination of convolutional and linear layers, thus endowing its features with equivariance and scalability. This enables the network to produce implicit representations for both single objects and large-scale scenes. 
Our implementation uses the network variant that stores features on a 3D regular grid. 

\vspace{0.05in}
\noindent\textbf{Encoder.}
The encoder is a shallow PointNet \cite{qi2017pointnet} but with local pooling layers. By using these input features generated by the local PointNet encoder, we obtain a $32^3$ volumetric feature grid that captures the local information in the neighborhood of the points, which is necessary to capture local geometric information about the shape of the input point cloud. Due to memory constraints, the volumetric feature can represent rich 3D information but is restricted to small resolutions and sizes.

\vspace{0.05in}
\noindent\textbf{Decoder.}
To endow the encoder features with inductive bias, the occupancy network uses a 3D UNet encoder \cite{iek20163DUL} to process the volumetric feature grid. Since U-Net contains convolutional operations, this also introduces translational equivariance in the encoder features, which makes it able to predict the occupancy of different shapes but from the same categories. These aggregated feature maps from the U-Net \cite{ron2015unet} are then fed into a decoder for predicting occupancy labels. To predict the occupancy value at any arbitrary position, we use tri-linear interpolation to find the features at that point by using the features of all points belonging to the same voxel in the volumetric grid. This point's location and features are passed through a decoder that outputs an occupancy value for each 3D grid location.

\vspace{0.05in}
\noindent\textbf{Surface Sampling.} 
We render an output mesh of the given input point cloud from the predicted occupancy of the grid points of the convolutional occupancy network using the MISE algorithm~\cite{mescheder2019occupancy}. We then produce an augmented version $\mathbf{x}_k$ of the original point cloud $\mathbf{x}_0$ by randomly sampling a point cloud from the vertices of the rendered mesh, where $k$ indicates the $k$-th augmentation.

\subsection{Augmentation by Point Cloud Upsampling}
\label{sec:self_up}
Inspired by \cite{sapcu}, we upsample input sparse point cloud $\mathbf{x} = \{\mathbf{p}_i\}_{i=1}^n \in \mathbb R^{n\times3}$ to dense point cloud $\mathbf{y} = \{\mathbf{p}_i\}_{i=1}^N\in \mathbb R^{N\times3}$ including $N = \lfloor r\times n \rfloor$ points, where $r$ is desirable scaling factor (set default to 4). The high-resolution point cloud $\mathbf{y}$ must be dense, uniform, complete, and noise-tolerant. The self-supervised point cloud upsampling strategy includes four steps: seeds sampling, surface projection, outliers removal, and arbitrary-scale point cloud generation. 

\vspace{0.05in}
\noindent\textbf{Seeds Sampling.} 
To obtain uniformly sampled seed points, given a point cloud, we divide the 3D space into equally spaced voxels and estimate the distance from centers to the surface by computing the distance to the triangles formed by the nearest points. Then we choose the centers in a preset range as the seed points.

\vspace{0.05in}
\noindent\textbf{Surface Projection.} 
Given a seed point $c$, we obtain the coordinate of the projection point of the seed point $c$ as: $c_p=c+n \times d$, where $\mathbf{n} \in[-1,1]^3$ and $d \in \mathbb{R}$ are projection direction and projection distance, respectively. The $\mathbf{n}$ and $d$ can be obtained by two multi-layer fully-connected neural networks $f_n$, and $f_d$, which borrows from Occupancy Network \cite{mescheder2019occupancy} and DGCNN \cite{wang2019dgcnn}. The detail of architectures and training procedures can be found in \cite{sapcu}. 

\vspace{0.05in}
\noindent\textbf{Outliers Removal.} 
For a projection point $c_p$, we determine a point as an outlier if $b_p>1.5 \overline{\mathrm{b}}$, where $b_p$ is the average bias between $c_p$ and its nearest points and $\overline{\mathrm{b}}$ is the average bias of all projection points. 

In practice, outlier removal can be regarded as optional, but we empirically found that outlier removal can yield some minor performance improvement of downstream tasks such as classification and part segmentation, and therefore use this step by default in the augmentation.

\vspace{0.05in}
\noindent\textbf{Point Cloud Generation.} 
We upsample the input point cloud $\mathbf{x}_0$ to a dense point cloud $\mathbf{y}$ using the upsampling network. Then, we sample a fixed number of points from the upsampled point cloud $\mathbf{y}$ by using the farthest-point sampling algorithm to obtain an augmented point cloud $\mathbf{x}_k$ with the desired number of points, where $k$ indicates the $k$-th augmented point cloud. The examples are shown in Figure~\ref{fig:recon_compare}.

\subsection{Downstream Tasks.}

\noindent\textbf{Object Classification.}
To leverage the augmented point clouds for classification, for both PointNet~\cite{qi2017pointnet}, DGCNN~\cite{wang2018edgeconv}, and PointNeXt~\cite{qian2022pointnext}, we extract the global features of each point cloud $\mathbf{x}_k$ including the original point cloud $\mathbf{x}_0$, and then take an average of the features before passing them to the classifier. Without changing of notation, assume that $f$ is the global feature extractor, and $g$ is the classifier, we can write the label prediction as:
\begin{align}
    g(\mathrm{avgpool}(f(\mathbf{x}_0), f(\mathbf{x}_1), f(\mathbf{x}_2), ...))
\end{align}

\noindent\textbf{Semantic and Part Segmentation.}
For semantic segmentation and part segmentation, the aggregation function is more evolved. The basic idea is first to perform segmentation on each point cloud, and then aggregate the results to produce the final segmentation for the original point cloud $\mathbf{x}_0$, but now the aggregation occurs at a per-point level instead of the global features. Let $f_i(\mathbf{x}_k)$ be the features of point $i$ in point cloud $\mathbf{x}_k$, the label prediction of point $i$ in the original point cloud $\mathbf{x}_0$ can be written as:
\begin{align}
    g(\phi( & f_i(\mathbf{x}_0), \{ f_{\pi_{1,i}}(\mathbf{x}_1) \}, \{ f_{\pi_{2,i}}(\mathbf{x}_2) \}, ...))
\end{align}
where $\pi_{k, i}$ indicates the corresponding points of point $i$ in $\mathbf{x}_0$ to point cloud $\mathbf{x}_k$, and $g$ as the classifier or any post-processing network. Here we propose a simple algorithm to establish such correspondences via nearest neighbors on the logit vectors, which are detailed in Algorithm \ref{alg:tta}.

\begin{algorithm}[!ht]
 	\caption{Pseudo-code for our test-time augmentation for the segmentation task.}
 	\label{alg:tta}
	
 	\definecolor{codeblue}{rgb}{0.28,0.52,0.76}
 	\definecolor{keywordorange}{rgb}{0.98, 0.51, 0.14}
 	\lstset{
 		basicstyle=\fontsize{8pt}{8pt}\ttfamily\bfseries,
 		commentstyle=\fontsize{7pt}{7pt}\color{codeblue},
 		keywordstyle=\bfseries\fontsize{11pt}{11pt}\color{keywordorange},
 	}
\begin{lstlisting}[language=python]
# get_log(p):  return logit at point p.
# get_feat(p): return 3D coords wo/ or w/ logit
# knn(p, X):   return nearest neighbor of p in X. 
# agg(a, b):   combine tensors a and b.
for each point p in X_0:
  logit = get_log(p)
  feat = get_feat(p)
  for i = 1 to N                  
    neighbors = knn(feat, X_i) 
    for each point q in neighbors:
      logit = agg(logit, get_log(q))
  label = argmax(logit) 
\end{lstlisting}
\end{algorithm}

\section{Experimental Results}
\label{sec:experiments}

\begin{table}[!t]
\centering
\caption{3D object classification in ModelNet40 \cite{wu20153d} and ScanObjectNN \cite{uy-scanobjectnn-iccv19} using self-supervised upsampling point clouds~\cite{sapcu}.}
\label{tab:classification}
\resizebox{0.46\textwidth}{!}{%
\begin{tabular}{@{}l|cc|cc@{}}
\toprule
\multicolumn{1}{c|}{\multirow{2}{*}{Method}} & \multicolumn{2}{c|}{\shortstack{\textbf{ModelNet40}\\~}}   & \multicolumn{2}{c}{\shortstack{\textbf{ScanObjectNN}\\ (PB\_T50\_RS)}} \\ \cmidrule(l){2-5} & oAcc & mAcc & oAcc & mAcc \\ \midrule
PointNet \cite{qi2017pointnet}         & 89.20                & 86.20              & 68.20              & 63.40                       \\
Ours                                   & \textbf{92.07}       & \textbf{88.78}     & \textbf{76.69}     & \textbf{72.93}              \\ \midrule
DGCNN \cite{wang2019dgcnn}             & 92.90                & 90.20              & 78.10              & 73.60                       \\
Ours                                   & \textbf{94.23}       & \textbf{91.79}     & \textbf{87.71}     & \textbf{85.84}              \\ \midrule
PointNeXt \cite{qian2022pointnext}     & 93.96                & 91.14              & 88.18              & 86.83                       \\
Ours                                   & \textbf{95.48}       & \textbf{92.96}     & \textbf{90.38}     & \textbf{88.99}              \\ \midrule
PointMixer \cite{pointmixer}           & 91.41                & 87.89              & 82.51              & 80.03                       \\ 
Ours                                   & \textbf{92.71}       & \textbf{90.42}     & \textbf{84.18}     & \textbf{81.25}              \\ \midrule 
PointTransformer \cite{zhao2021point}  & 90.64                & 87.84              & 82.31              & 80.77                       \\ 
Ours                                   & \textbf{92.55}       & \textbf{89.73}     & \textbf{83.66}     & \textbf{81.37}              \\
\bottomrule
\end{tabular}%
}
\end{table}

\subsection{Implementation Details}

\begin{table*}[!htb]
\small
\begin{minipage}[t]{.44\linewidth}
\centering
\caption{Part segmentation on ShapeNet~\cite{chang2015shapenet} using self-supervised upsampling point clouds as input. }
\label{tab:shapenetpart_upsampling}
\resizebox{0.66\textwidth}{!}{%
\begin{tabular}{l|cc}
\toprule
2048 points    & mInsIoU & mCatIoU \\ 
\midrule
PointNet  \cite{qi2017pointnet}  & 80.74                       & 83.73 \\
Ours                             & \textbf{82.88}              & \textbf{86.25} \\ 
\midrule
DGCNN \cite{wang2019dgcnn}       & 81.08                       & 84.18 \\
Ours                             & \textbf{83.38}              & \textbf{86.70} \\ 
\midrule
PointNeXt \cite{qian2022pointnext} & 84.23                     & 86.73 \\
Ours                             & \textbf{85.07}              & \textbf{87.60} \\ 
\bottomrule
\end{tabular}%
}
\end{minipage}%
\hfill%
\begin{minipage}[t]{.50\linewidth}
\centering
\caption{Part segmentation on ShapeNet~\cite{chang2015shapenet} using surface sampling with different numbers of points. } 
\label{tab:part_pointnet_shapenet}
\resizebox{0.96\textwidth}{!}{%
\begin{tabular}{l|cc|cc}
\toprule
\multicolumn{1}{c|}{\multirow{2}{*}{\textbf{Method}}}  & \multicolumn{2}{c|}{\textbf{128 points}} & \multicolumn{2}{c}{\textbf{256 points}} \\ \cmidrule(l){2-5} 
& mInsIoU & mCatIoU & mInsIoU & mCatIoU  \\ \midrule
PointNet  \cite{qi2017pointnet} & 79.06 & 81.72 & 83.12 & 85.12 \\
Ours & \textbf{79.55} & \textbf{82.66} & \textbf{83.25} & \textbf{85.82} \\
\midrule 
DGCNN \cite{wang2019dgcnn} & 59.75 & 66.34 & 69.88 & 74.57 \\
Ours & \textbf{71.63} & \textbf{81.95} & \textbf{79.98} & \textbf{85.65} \\ 
\bottomrule
\end{tabular}%
}
\end{minipage} 
\end{table*}

\begin{table}[t]
\small
\centering
\caption{Semantic segmentation on SemanticKITTI \cite{behley2019semantickitti} using self-supervised upsampling point clouds. }
\label{tab:semantickitti}
\begin{tabular}{@{}l|cc@{}}
\toprule
Method           & mAcc & mIoU \\ \midrule
RandLANet \cite{hu2020randla}                & 97.23              & 68.84              \\
Ours & \textbf{99.17}              & \textbf{70.55}              \\ \bottomrule
\end{tabular}
\end{table}

\begin{table}[t!]
\centering
\caption{Ablation studies of our test-time augmentation on ShapeNet~\cite{chang2015shapenet} using surface sampling. Performing k-nearest neighbor search on high-dimensional feature space (model A, B) and using the average function (model B) for aggregating predictions result in improved accuracies. The performance can be further boosted by using extra augmented point clouds (model A\&C and B\&C). The reported metric is mCatIoU.}
\label{tab:analysis}

\resizebox{0.42\textwidth}{!}{%
\begin{tabular}{l|cc}
\toprule
2048 points & PointNet~\cite{qi2017pointnet}  & DGCNN~\cite{wang2019dgcnn} \\ \midrule
xyz (max)         & 86.45 & 84.26 \\
A: w/ logit (max) & 88.30 & 85.90 \\
B: w/ logit (avg) & 88.43 & 86.05 \\
C: w/ 10x samples & 86.39 & 84.13 \\
A\&C (max)        & 88.26 & 85.96 \\
B\&C (avg)        & \textbf{88.58} & \textbf{86.16} \\
\bottomrule
\end{tabular}%
}
\end{table}

We implement our method in Pytorch. We use the convolutional occupancy network~\cite{peng2020convolutional}, and self-supervised point upsampling network~\cite{sapcu} for test-time augmentation. For downstream tasks, we experiment with pre-trained models for classification and part segmentation such as PointNet~\cite{qi2017pointnet}, DGCNN~\cite{wang2019dgcnn}, PointNeXt \cite{qian2022pointnext} as well as for large-scale semantic scene segmentation such as RandLANet \cite{hu2020randla}.

\vspace{0.05in}
\noindent\textbf{Dataset and metric.}
Our experiments are conducted on different datasets such as ShapeNet~\cite{chang2015shapenet}, ScanObjectNN \cite{uy-scanobjectnn-iccv19},  ModelNet40~\cite{wu20153d}, and SemanticKITTI~\cite{behley2019semantickitti} datasets, including indoor and outdoor environments with both synthetic and real data. the ShapeNet dataset~\cite{chang2015shapenet}.

We employ several popular metrics for evaluation, such as the overall and mean percentage accuracy are computed for the classification task, the Instance and Category Intersection-Over-Union (mInsIoU, mCatIoU) are utilized for the part segmentation task, and the mean Accuracy (mACC) and mean IoU (mIoU) are used for semantic segmentation task.

\vspace{0.05in}
\noindent\textbf{Data processing.}
For ShapeNet~\cite{chang2015shapenet} dataset, we use the pre-processed data produced by PointNet~\cite{qi2017pointnet}, which is an early version of ShapeNet (version 0) to train the segmentation network. Nonetheless, the ShapeNet data used to train the convolutional occupancy network~\cite{peng2020convolutional} is a different version (version 1). Since the number of objects differs in these variants of ShapeNet, we only use the objects that appear in both datasets. For ScanObjectNN \cite{uy-scanobjectnn-iccv19}, and ModelNet40~\cite{wu20153d} datasets, we follow the instruction in the official implementation of PointNeXt \cite{qian2022pointnext}.
For SemanticKITTI~\cite{behley2019semantickitti} dataset, we follow the instruction in the official implementation of RandLANet \cite{hu2020randla}. We also follow Self-UP \cite{sapcu} to prepare the data for point cloud upsampling.

\begin{figure}[t]
    \centering
    \def\sc{0.49}
    \includegraphics[width=\sc\linewidth]{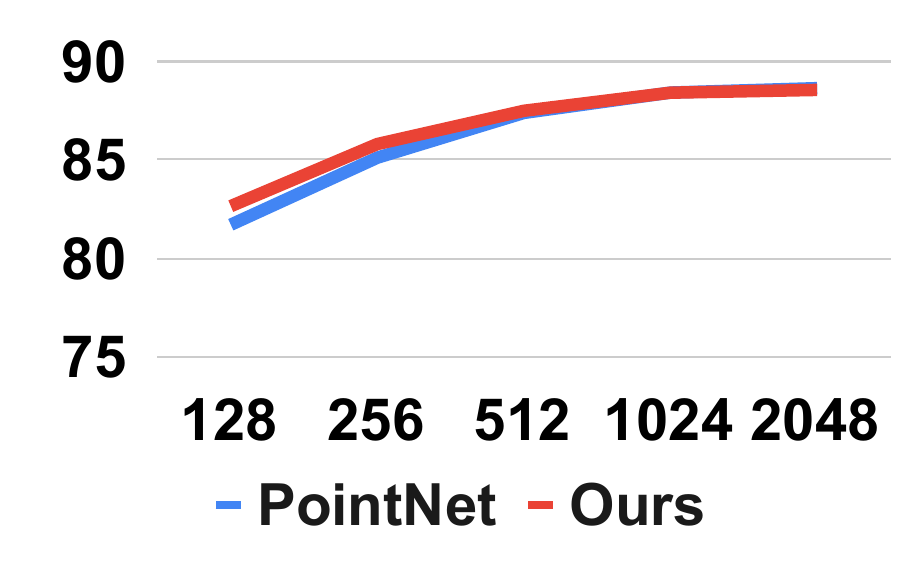}
    \includegraphics[width=\sc\linewidth]{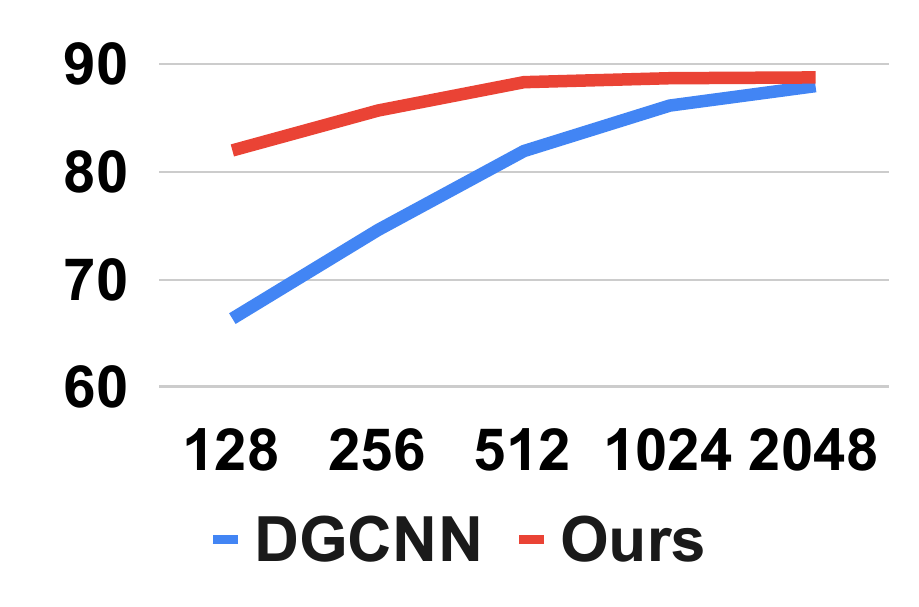}
    \\
    \includegraphics[width=\sc\linewidth]{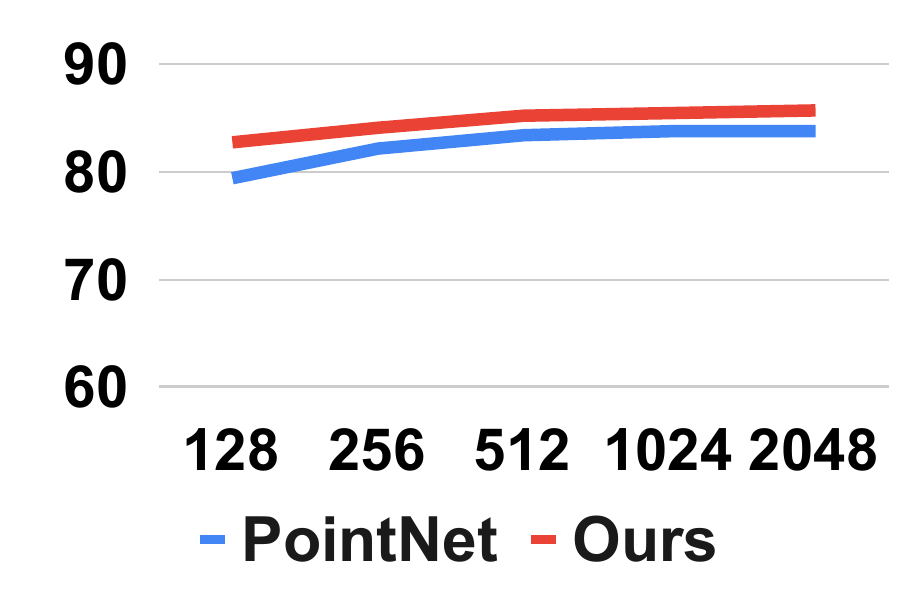}
    \includegraphics[width=\sc\linewidth]{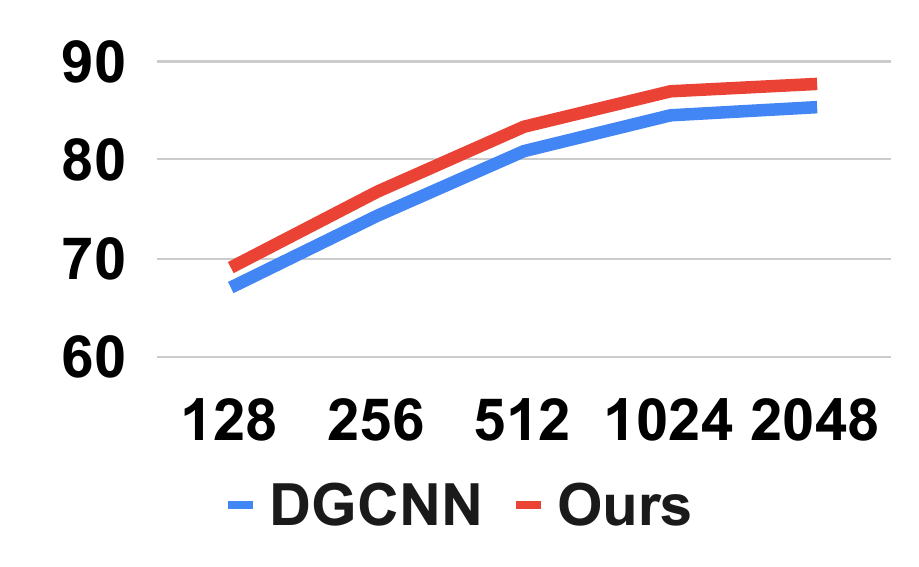}
    \caption{TTA using surface sampling (top row) and self-supervised upsampling (bottom row) on part segmentation on ShapeNet with different numbers of points. We found that applying TTA for sparse point clouds of the surface sampling method yields significant improvement. In contrast, the improvement of TTA on upsampling point clouds is more stable, thanks to dense and uniformly distributed proximity-to-surface point clouds. The horizontal axis is in log scale. Best viewed with zoom. }
    \label{fig:plot_points}
\end{figure}

\vspace{-0.in}
\subsection{Classification Results}
The object classification results are shown in Table~\ref{tab:classification} and are conducted on two challenging datasets (ScanObjectNN \cite{uy-scanobjectnn-iccv19}, and ModelNet40~\cite{wu20153d}). ScanObjectNN presents considerable problems to the various point cloud analysis algorithms already in use due to occlusions and noise. Based on PointNeXt \cite{qian2022pointnext}, we conduct experiments on PB\_T50\_RS, the most challenging and widely deployed version of ScanObjectNN. Note that the reported performance of PointNet and DGCNN in our paper is higher than the original PointNet and DGCNN paper because we adopt the re-implementation of PointNet and DGCNN from the PointNeXt paper, which includes optimized training strategies. 

As can be seen, the optimized baseline model by PointNet~\cite{qi2017pointnet} performed very well on both ModelNet40 and ScanObjectNN classification. Despite such, applying augmentation with our method leads to a performance boost of $1-2\%$, which is a significant gain given the saturating accuracy of this dataset. We also empirically found that augmenting with more than one sampled point cloud does not significantly improve this task. 

Note that as convolutional occupancy network~\cite{peng2020convolutional} requires ground truth signed distance functions to train surface reconstruction, we only perform the classification task using self-supervised point upsampling~\cite{sapcu}.

\subsection{Segmentation Results}

\begin{table*}[!htb]
\begin{minipage}[t]{.35\linewidth}
\centering
\caption{Comparison of different augmentation methods on part segmentation with PointNet \cite{qi2017pointnet} as backbone on ShapeNet~\cite{chang2015shapenet} dataset. TTA is done using Screened Poisson reconstruction~\cite{Kazhdan2006}, Neural Implicit representation \cite{peng2020convolutional}, and Point Clouds Upsampling~\cite{sapcu}.} 
\label{tab:comparison_recon}
\begin{tabular}{l|c|c}
\toprule
128 points & mCatIoU  & mInsIoU \\ \midrule
Poisson \cite{Kazhdan2006} & 81.79 &  77.86\\
Implicit \cite{peng2020convolutional} & 82.66 &  79.55\\
Self-UP \cite{sapcu} & 82.70 & 78.99\\
\bottomrule
\end{tabular}%
\end{minipage}%
\hfill%
\begin{minipage}[t]{.28\linewidth}
\centering
\caption{Comparison with traditional augmentation with different values of $\sigma$ on the classification task using surface sampling on ShapeNet~\cite{chang2015shapenet}. The backbone of our TTA is PointNet~\cite{qi2017pointnet}.}
\label{tab:noise}
\begin{tabular}{@{}l|l}
\toprule
2048 points & mAcc \\ \midrule
$\sigma = 0.05$ & 98.21 \\
$\sigma = 0.07$ & 97.69  \\
$\sigma = 0.1$ & 96.54 \\
Ours & \textbf{98.53}  \\
\bottomrule
\end{tabular}%
\end{minipage}%
\hfill%
\begin{minipage}[t]{.33\linewidth}
\centering
\caption{Augmentation with normals using surface sampling on ShapeNet~\cite{chang2015shapenet}: we classify an original point cloud w/o normal vectors by using an implicit field to sample the normals and then classify the augmented point cloud. The backbone is PointNet~\cite{qi2017pointnet}.} 
\label{tab:augmentation_normals}
\begin{tabular}{@{}l|l}
\toprule
2048 points & mAcc \\ \midrule
Org. xyz & 97.73 \\
Aug. xyz & 98.53  \\
Aug. xyz \& normals & 98.38  \\
\bottomrule
\end{tabular}%
\end{minipage} 
\end{table*}

\begin{figure}[!ht]
\centering
\def\sc{0.186}

\includegraphics[width=\sc\linewidth]{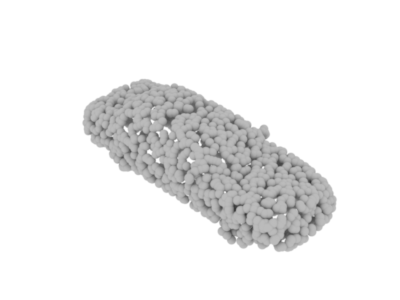}
\includegraphics[width=\sc\linewidth]{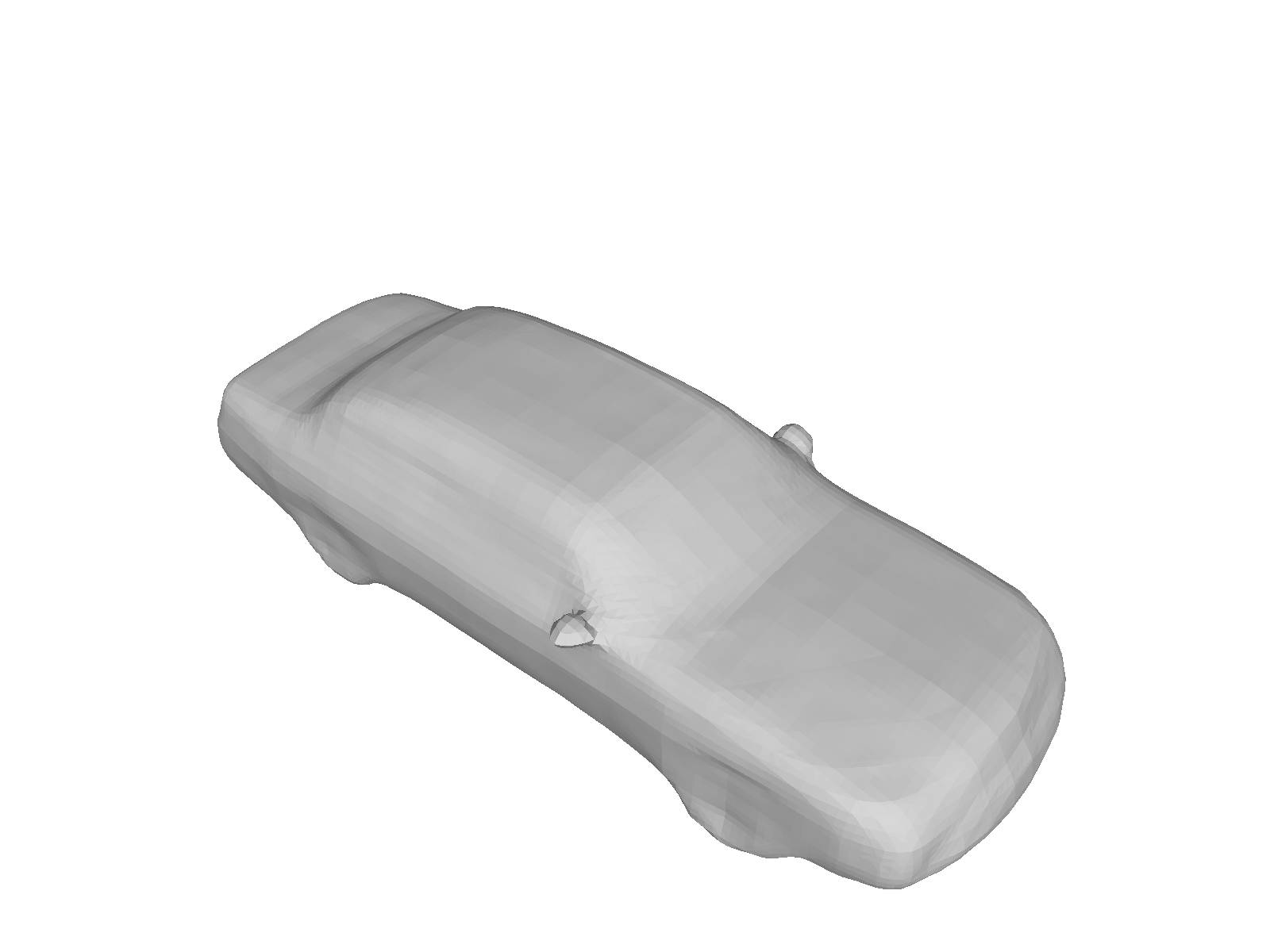}
\includegraphics[width=\sc\linewidth]{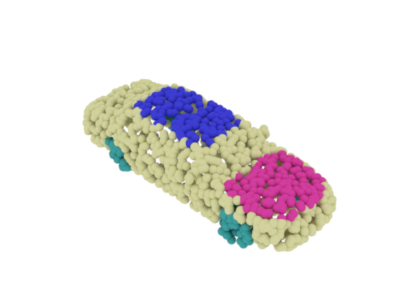}
\includegraphics[width=\sc\linewidth]{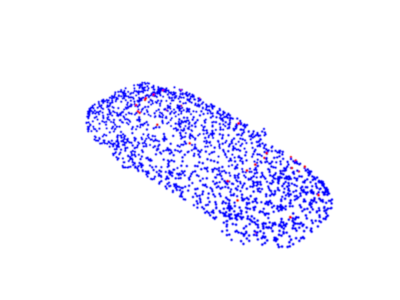}
\includegraphics[width=\sc\linewidth]{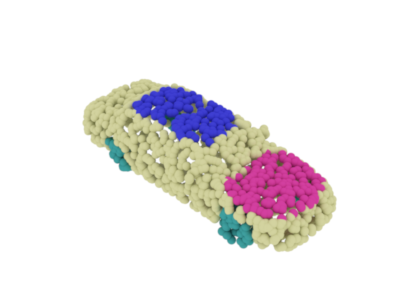}

\includegraphics[width=\sc\linewidth]{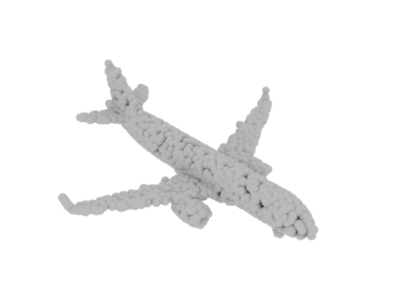}
\includegraphics[width=\sc\linewidth]{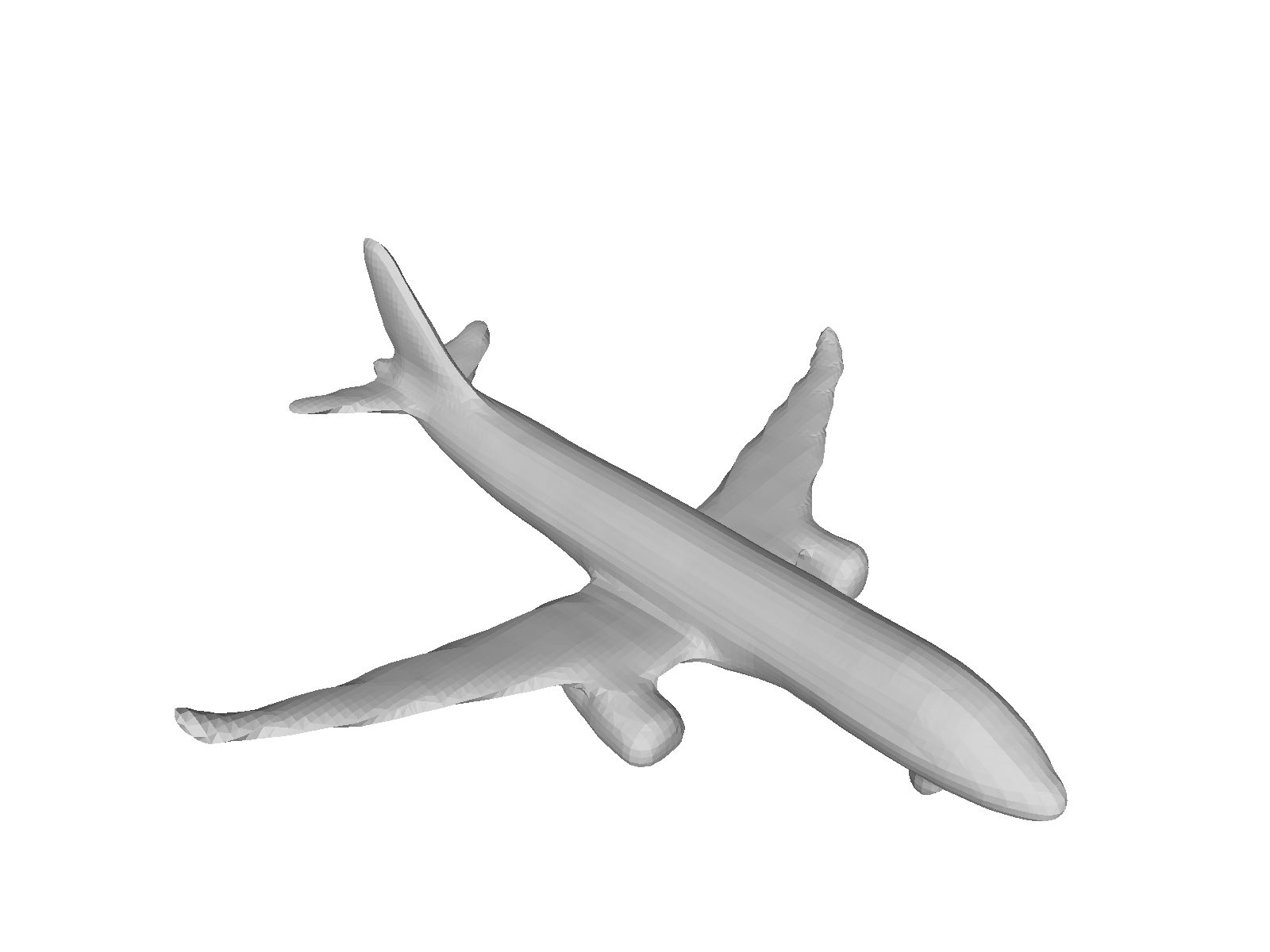}
\includegraphics[width=\sc\linewidth]{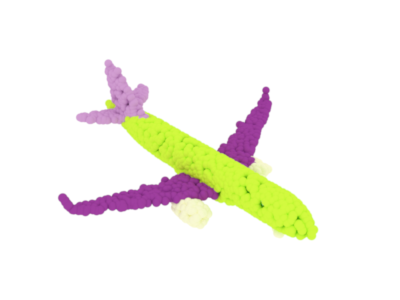}
\includegraphics[width=\sc\linewidth]{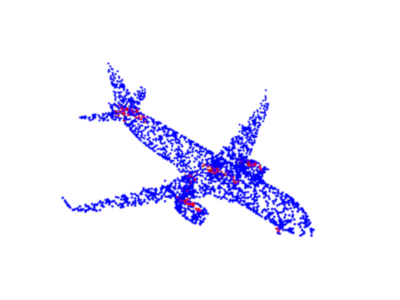}
\includegraphics[width=\sc\linewidth]{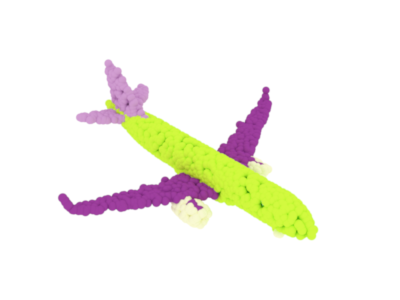}

\includegraphics[width=\sc\linewidth]{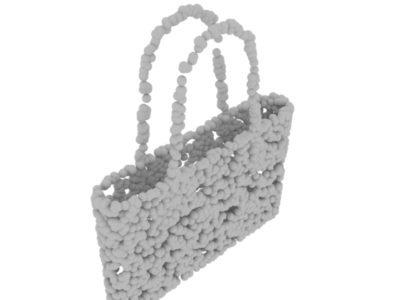}
\includegraphics[width=\sc\linewidth]{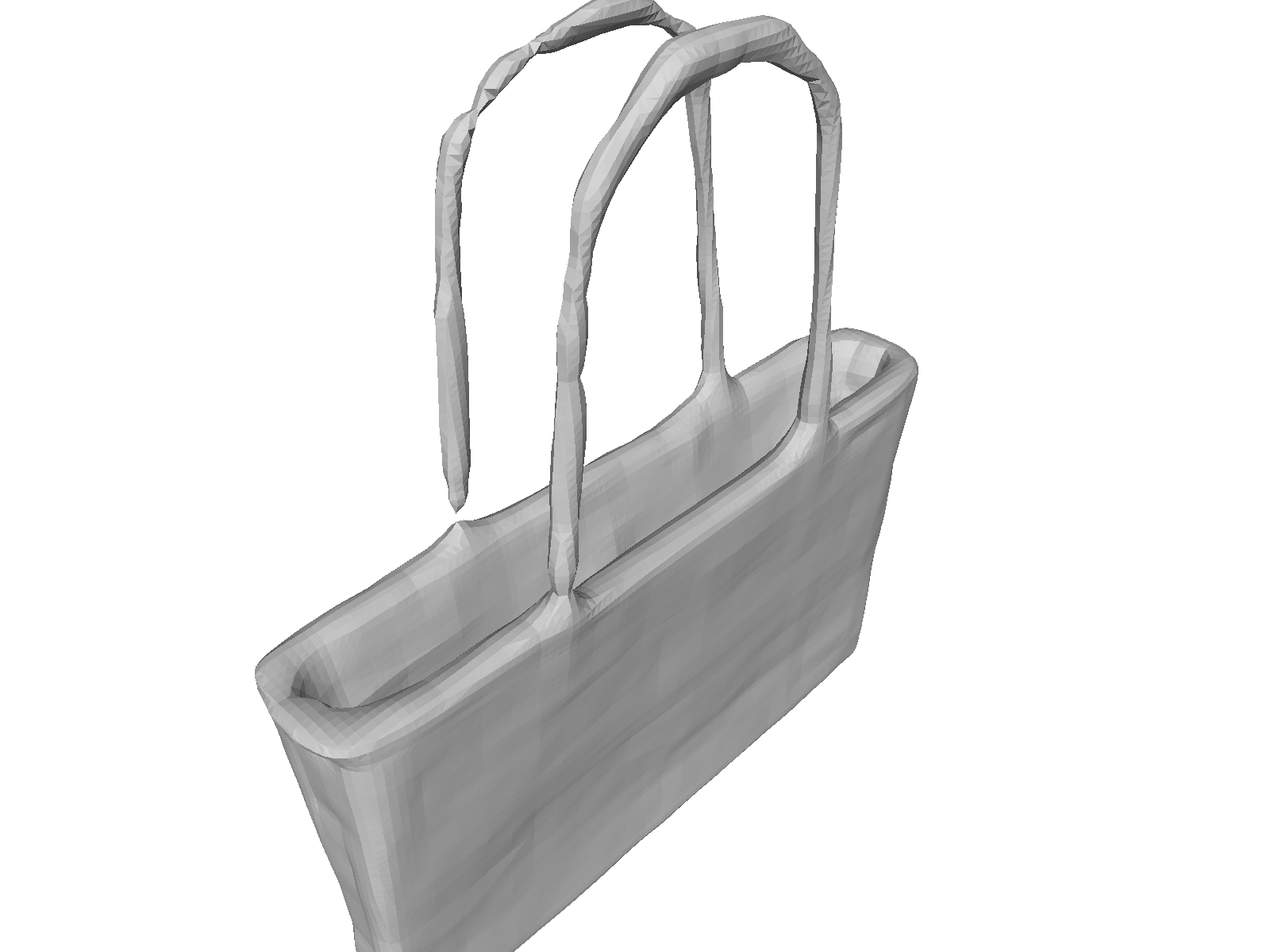}
\includegraphics[width=\sc\linewidth]{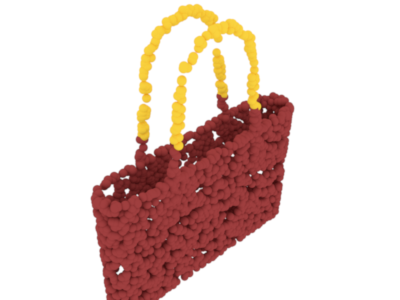}
\includegraphics[width=\sc\linewidth]{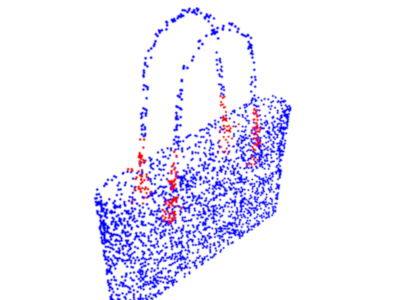}
\includegraphics[width=\sc\linewidth]{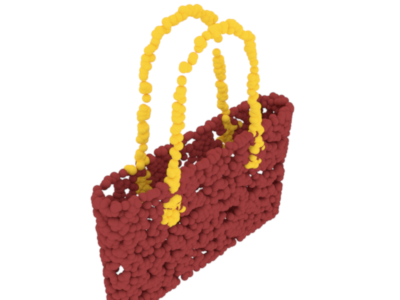}

\includegraphics[width=\sc\linewidth]{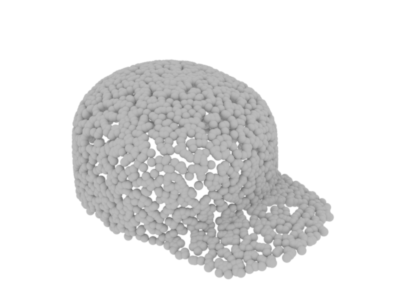}
\includegraphics[width=\sc\linewidth]{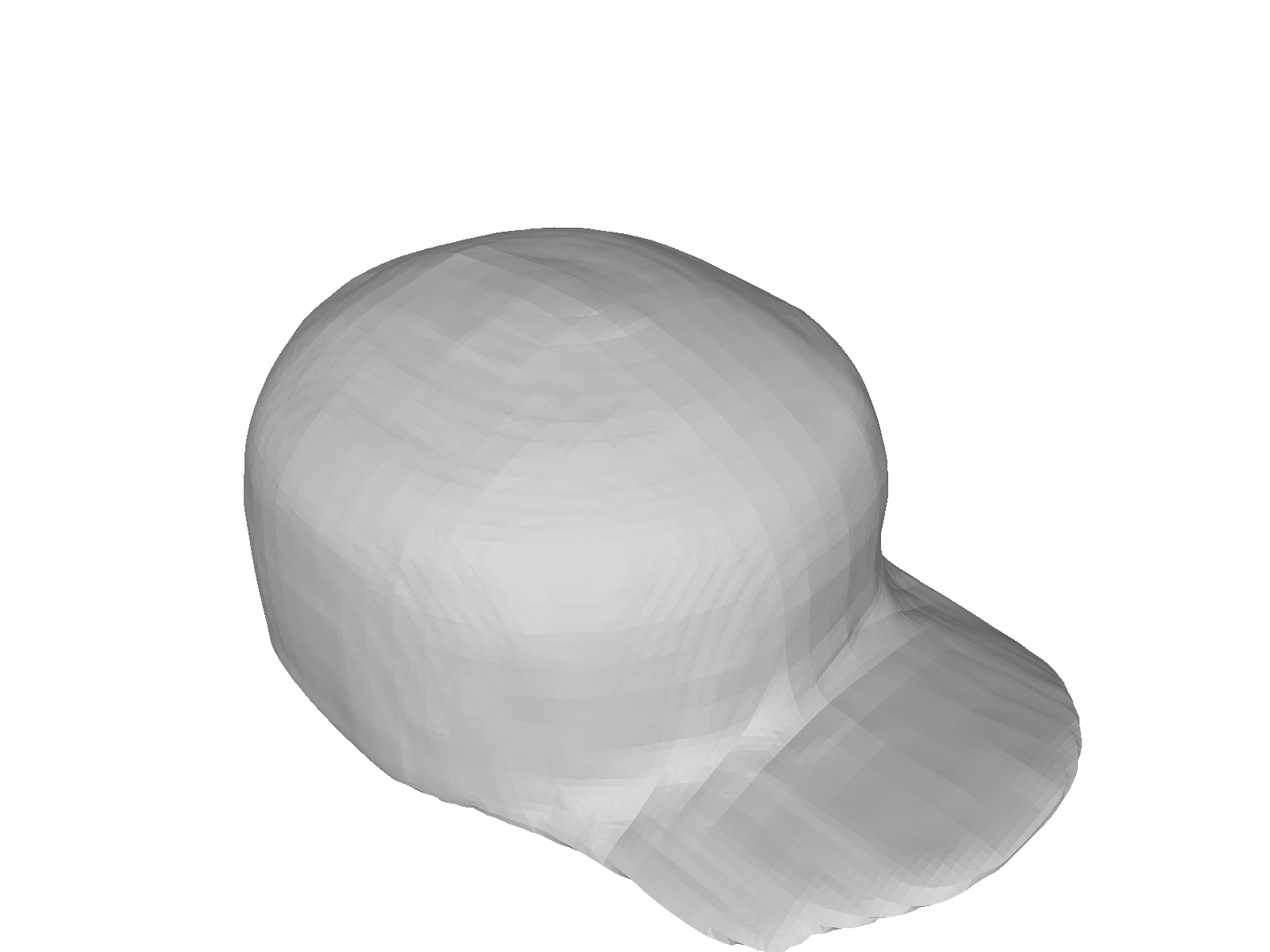}
\includegraphics[width=\sc\linewidth]{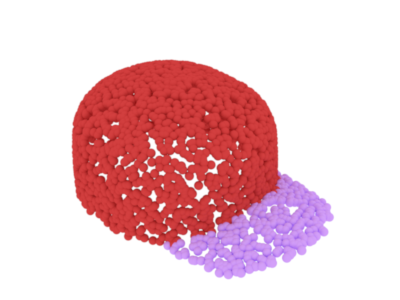}
\includegraphics[width=\sc\linewidth]{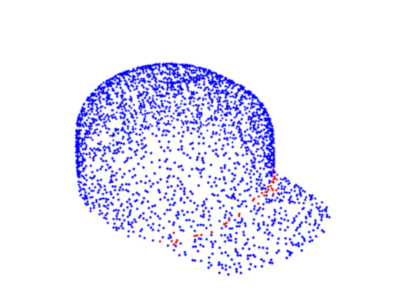}
\includegraphics[width=\sc\linewidth]{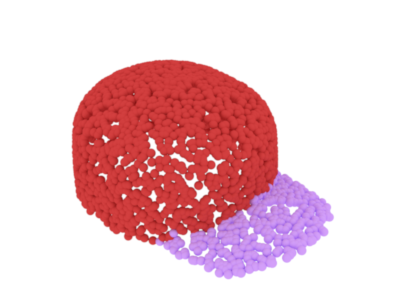}

\includegraphics[width=\sc\linewidth]{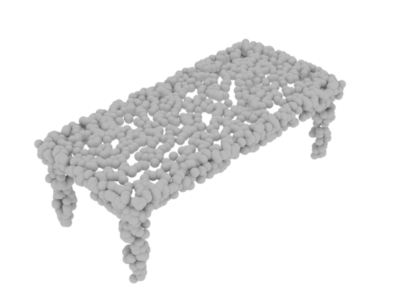}
\includegraphics[width=\sc\linewidth]{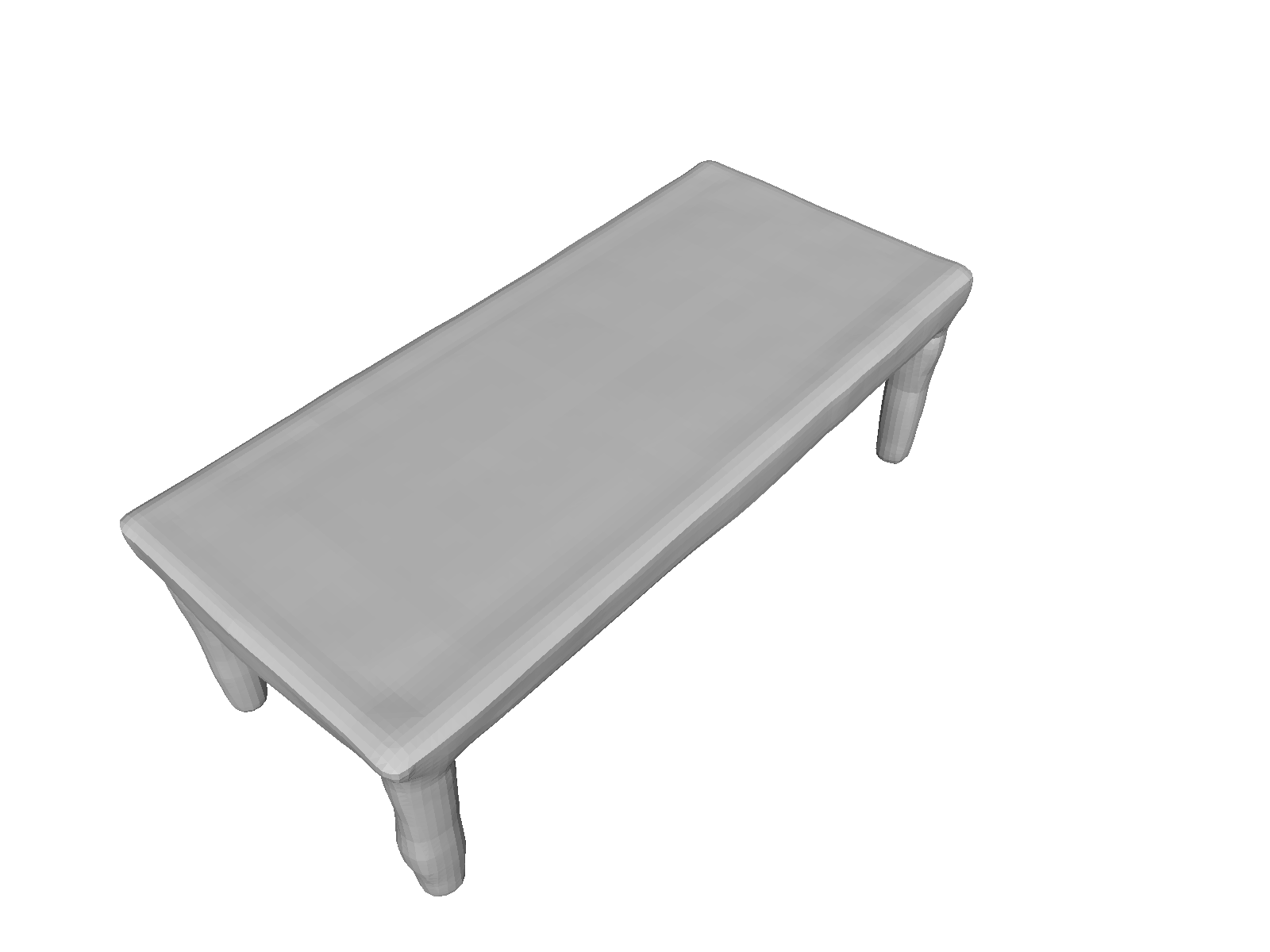}
\includegraphics[width=\sc\linewidth]{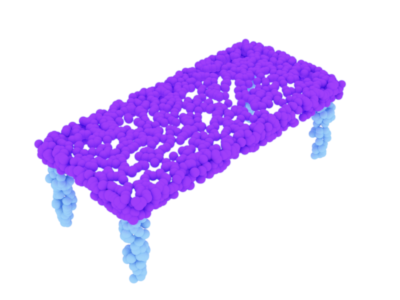}
\includegraphics[width=\sc\linewidth]{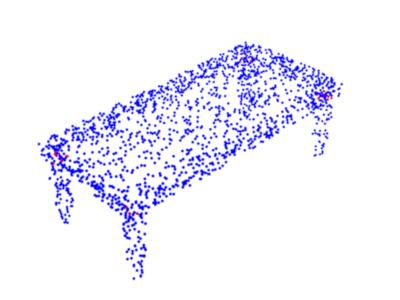}
\includegraphics[width=\sc\linewidth]{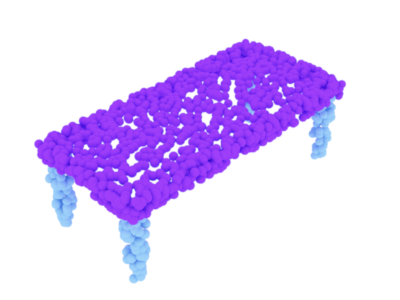}

\vspace{0.05in}
\caption{\textbf{Visualization of part segmentation results.} From left to right: Input, Reconstruction, Ours, Difference map, and GT. In the difference map, where blue and red points indicate correct and wrong labels, respectively, our test-time augmentation mainly deals with the labels along the boundaries, improving accuracies through aggregating predictions from augmented point clouds.}
\label{fig:part_seg_point}
\end{figure}

\noindent\textbf{Part Segmentation.}
The part segmentation results are shown in Table~\ref{tab:shapenetpart_upsampling} and Table~\ref{tab:part_pointnet_shapenet}. It can be seen that by applying our method, the mInsIoU, and mCatIoU are improved compared to the baseline approach. The results also demonstrate the robustness of our method as it works well with different network backbones, e.g., PointNet~\cite{qi2017pointnet} that involves only per-point and global point cloud features, DGCNN~\cite{wang2018edgeconv} which establishes and learns dynamic graphs in point neighborhoods, and the SOTA PointNeXt \cite{qian2022pointnext}. It is worth noting that the improvement is mainly gained from the refinement of the segmentation boundaries (Figure~\ref{fig:part_seg_point}).

\vspace{0.05in}
\noindent\textbf{Semantic Segmentation.}
To assess the generalizability of our strategy, we also tested on real-world data from SemanticKITTI~\cite{behley2019semantickitti} dataset. As SemanticKITTI data is captured by LiDAR sensors, it is favorable to use point upsampling as the augmentation technique. It can be seen in Table \ref{tab:semantickitti}, by applying our method, the mAcc and mIoU are improved compared to the baseline approach.

\subsection{Additional Analysis} 

We perform additional experiments to validate the performance of our test-time augmentation. We select the segmentation task for these experiments as it produces dense prediction, which can be seen as generalized classifications. 

\vspace{0.05in}
\noindent\textbf{Point density.}
In Figure~\ref{fig:plot_points}, we plot the segmentation accuracies (mIoU) across different numbers of input points. Specifically, we randomly sample 128, 256, 512, 1024, and 2048 points as input to perform the segmentation. Compared to PointNet, it can be seen that our augmentation offers  significant performance improvement on sparse point clouds (128 and 256 points) and performs similarly to PointNet when the input points get denser.

We also found that by varying the number of input points (Figure~\ref{fig:plot_points}), DGCNN cannot perform well on sparse point clouds with a very large performance gap between the sparse and dense point clouds (more than 20\% between 128 and 2048 points). This is because for sparse point clouds, the neighbor graphs by DGCNN degenerate~\cite{wang2019dgcnn}. Despite such, our test-time augmentation can still improve the performance and significantly reduce the performance gap to around 6\%. This shows that our test-time augmentation is robust to the number of input points.

\vspace{0.05in}
\noindent\textbf{Ablation study.}
We conduct an ablation study on the part segmentation task on ShapeNet and provide the results in Table~\ref{tab:analysis}. We select the segmentation task as it is a generalized form of classification at per-point level, and also aim to justify the more complex design choices in the aggregation function for this task. We use inputs with 2048 points. Our baseline is an implementation that k-nearest neighbors are performed with just 3D coordinates as features. By adding logits as features, we can have 2\% gain in mIoU (model A). We also test different aggregate functions like max pooling and average pooling, and find that average pooling performs better (model B vs. C). Additionally, we repeat the sampling to obtain multiple augmented point clouds. By fusing the segmentation of these augmented point clouds to the original point cloud, further improvement can be achieved (model A\&C and B\&C). This shows that it is critical to compute accurate correspondences between the augmented point cloud and the original point cloud to achieve higher accuracies. 
From the above analysis, we can see that multi-sampling and changing aggregate functions can yield further improvement.

\vspace{0.05in}
\noindent\textbf{Comparison among augmentation techniques.} 
We provide a comparison to study which augmentation technique should be used in practice. We compare the popular Screened Poisson reconstruction~\cite{Kazhdan2006} to convolutional neural network \cite{peng2020convolutional} and point cloud upsampling~\cite{sapcu}. The results in Table \ref{tab:comparison_recon} show that our augmentation techniques are more favorable in performance than Screened Poisson reconstruction. The performance between the convolutional occupancy network and self-supervised upsampling are rather similar, with the convolutional occupancy network is slightly better in the instance IoU metric. We hypothesize that when (ground truth) surface information is available, it could be used to supervise the augmentation, else point cloud upsampling could be an effective and robust augmentation in several scenarios. We also explore a recent unsupervised reconstruction~\cite{Gropp2020} but find that the shape quality is poor compared to Screened Poisson reconstruction, and thus unsuitable for augmentation. Exploring more robust reconstruction could lead to interesting augmentation techniques for future work.

\vspace{0.05in}
\noindent\textbf{Comparison to traditional augmentation.}
Adding Gaussian noise is a commonly used traditional data augmentation scheme that perturbs the points by sampling from a Gaussian distribution. Combining the results from this perturbation in test-time augmentation is known as voting~\cite{klokov2017escape,liu2019relation}. In our implementation, we sample offsets from a zero-mean Gaussian with different standard deviation $\sigma$ and add the offsets back to the original point clouds to form augmented point clouds. 
For our TTA, we sample one more point cloud and then average the global features of the additional point cloud and the original point cloud before passing them to the classifier. As can be seen in Table~\ref{tab:noise}, our method outperforms the traditional augmentation scheme due to the implicit representation that allows more effective point sampling. Comparison with train-time augmentation is left for future work, as it requires model retraining, which is both more expensive and less robust especially when only pretrained models are provided.

\vspace{0.05in}
\noindent\textbf{Augmentation without normals.} 
We experiment with implicit surface representation for data augmentation in a practical setting where normals are not available. In this case, existing methods have to rely on the pure 3D coordinates (xyz), causing a performance decrease, while our method can easily solve this problem by sampling the 3D points as well as the normal vectors directly from the implicit surface. In this way, our method can maintain the performance to the same level regardless of the normal existence. This is verified in Table~\ref{tab:augmentation_normals} that our augmentation outperforms the baseline when only the 3D coordinates (xyz) are available.

\vspace{0.05in}
\noindent\textbf{Computation overhead.} 
While having better performance, modern TTA, including our method, relies on a neural network to predict augmented samples from each input and thus has more overhead compared to traditional methods. For example, if we use $M$ augmented point clouds, the overhead is approximate $M$ times the original time cost. To circumvent this problem, we propose to exploit parallelism and use batched inference instead. More discussions can be found in the supplementary material.

\section{Discussion and Conclusions}
\label{sec:conclusion}

We presented a new method for augmenting point clouds at test time by leveraging neural implicit and point upsampling networks to sample augmented point clouds and showed that such augmentation works effectively for the classification and semantic segmentation task. Our results are encouraging since this is one of the first attempts to design a test-time augmentation technique for 3D point cloud deep learning. 

A main difference between our TTA and traditional methods is that traditional methods only use simple transformations and are thus lightweight, but not input-aware and less robust. While our TTA requires more resources, the extra computation remains affordable and our method shows good results across tasks and datasets. We believe further explorations to reduce such performance trade-offs would be valuable contributions to this less-explored area of test-time augmentation for 3D point clouds.

\vspace{0.05in}
\noindent\textbf{Acknowledgment.} This research was supported by the Singapore Ministry of Education (MOE) Academic Research Fund (AcRF) Tier 1 grant (MSS23C010), and Ningbo 2025 Science and Technology Innovation Major Project (No. 2022Z072), and an internal grant from HKUST (R9429). This work is partially done when Srinjay Sarkar was a research resident at VinAI Research, Vietnam.

{
    \small
    \bibliographystyle{ieeenat_fullname}
    \bibliography{main}
}

\clearpage
\setcounter{page}{1}
\maketitlesupplementary

\section{Experiments}

\subsection{Computation overhead} 

While having better performance, modern TTA, including our method, relies on a neural network to predict augmented samples from each input and thus has more overhead compared to traditional methods. For example, if we use $M$ augmented point clouds, the overhead is approximately $M$ times the original time cost. To circumvent this problem, we propose to exploit parallelism and use batched inference instead. First, our computation overhead is sub-linear, which means that even if we have 10 times as many augmented samples (the same number of augmented samples that we used for all of our experiments), the overhead will only increase by a factor of two (see Table~\ref{tab:breakdown}). This overhead is manageable and can be reduced even further through the utilization of batched prediction, as demonstrated in Table ~\ref{tab:breakdown} below. Additionally, additional engineering like deploying the network to an inference-only framework (TensorFlow Lite) would further optimize inference. Second, we can reduce the number of augmented samples ($M \in \{2, 4, 8, 10\}$), which would result in milder improvement in comparison to the baseline but would incur significantly less overhead (see Table~\ref{tab:breakdown}). Finally, the overhead of TTA can be offset by its ease of use compared to other methods for performance improvement, e.g., when only pre-trained models are given or when retraining the entire model is not possible.

\begin{table}[!ht]
\centering
\caption{Running time breakdown of different stages of our TTA with different augmentation samples $M$ for the classification task on ModelNet40.}
\renewcommand\arraystretch{1.2}
\label{tab:breakdown}
\resizebox{\linewidth}{!}{%
\begin{tabular}{@{}c|ccccc@{}}
\toprule
\multicolumn{1}{c|}{}                    & M=10   & M=8    & M=4    & M=2    & PointNeXt               \\ \midrule
\midrule
\multicolumn{1}{c|}{B. forward 10x} & 0.6573 & 0.1025 & 0.1021 & 0.1016 & 0.2147 \\
B. GPU FPS 10x                      & 0.1036 & 0.5258 & 0.2892 & 0.1446 & 0.0                         \\
Aggregation                              & 0.0968 & 0.0959 & 0.0948 & 0.0934 & 0.0                        \\
Others                                   & 0.4382 & 0.4334 & 0.4295 & 0.4252 & 0.4382                  \\ \midrule
Total time                                   & 1.2959 & 1.1576 & 0.9156 & 0.7648 & 0.6529                  \\ \midrule
oAcc & 95.84 & 95.28 & 94.81 & 94.29 & 93.96 \\
mAcc & 92.96 & 92.51 & 91.87 & 91.38 & 91.14 \\
\bottomrule
\end{tabular}%
}
\end{table}

\subsection{Semantic Segmentation on S3DIS dataset}
\begin{figure*}[t]
    \centering
    \includegraphics[width=0.85\linewidth]{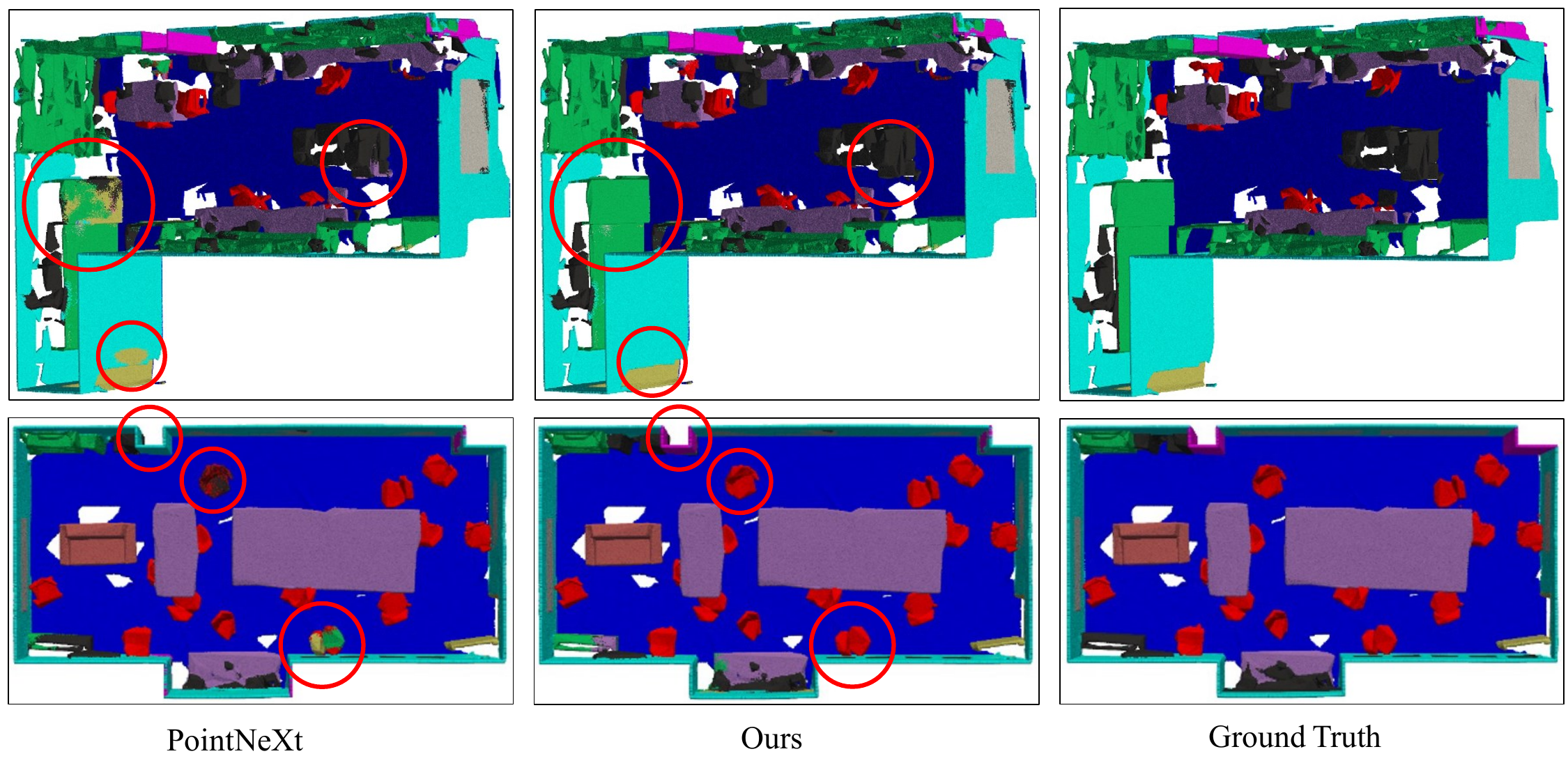}
    \caption{Semantic segmentation results of PointNeXt~\cite{qian2022pointnext}, ours and ground truth on S3DIS dataset~\cite{armeni20163d}.}
    \label{fig:visual_s3dis}
\end{figure*}

We provide quantitative and qualitative results of semantic segmentation on S3DIS dataset in Table~\ref{tab:s3dis} and Figure~\ref{fig:visual_s3dis}.
As can be seen, our TTA is effective and improves upon the baseline PointNeXt.

\begin{table*}[!ht]
\centering
\caption{TTA with point cloud upsampling~\cite{sapcu} on semantic segmentation on S3DIS~\cite{armeni20163d}.}
\label{tab:s3dis}
\resizebox{\textwidth}{!}{%
\begin{tabular}{@{}c|cc|ccccccccccccc@{}}
\toprule
\multirow{2}{*}{\textbf{Method}} & \multirow{2}{*}{\textbf{mIoU} (\%)} & \multirow{2}{*}{\textbf{\textbf{mAcc}} (\%)} & \multicolumn{13}{c}{\textbf{Category IoU}}                                                                    \\ \cmidrule(l){4-16} 
                        &                            &                            & ceiling & floor & wall  & beam & column & window & door  & table & chair & sofa  & bookcase & board & clutter \\ \midrule
PointNeXt               & 64.26                      & 70.69                      & 94.07   & 98.26 & 80.89 & 0    & 23.84  & 48.66  & 66.58 & 80.99 & 90.06 & 67.94 & 72.08    & 58.06 & 54.04   \\
Ours                    & \textbf{65.23}                      & \textbf{71.75}                      & 95.48   & 99.73 & 82.11 & 0    & 24.21  & 49.39  & 67.58 & 82.21 & 91.41 & 68.96 & 73.16    & 58.93 & 54.85   \\ \bottomrule
\end{tabular}
}
\end{table*}

\begin{figure*}[h!]
    \centering
    \includegraphics[width=0.85\linewidth]{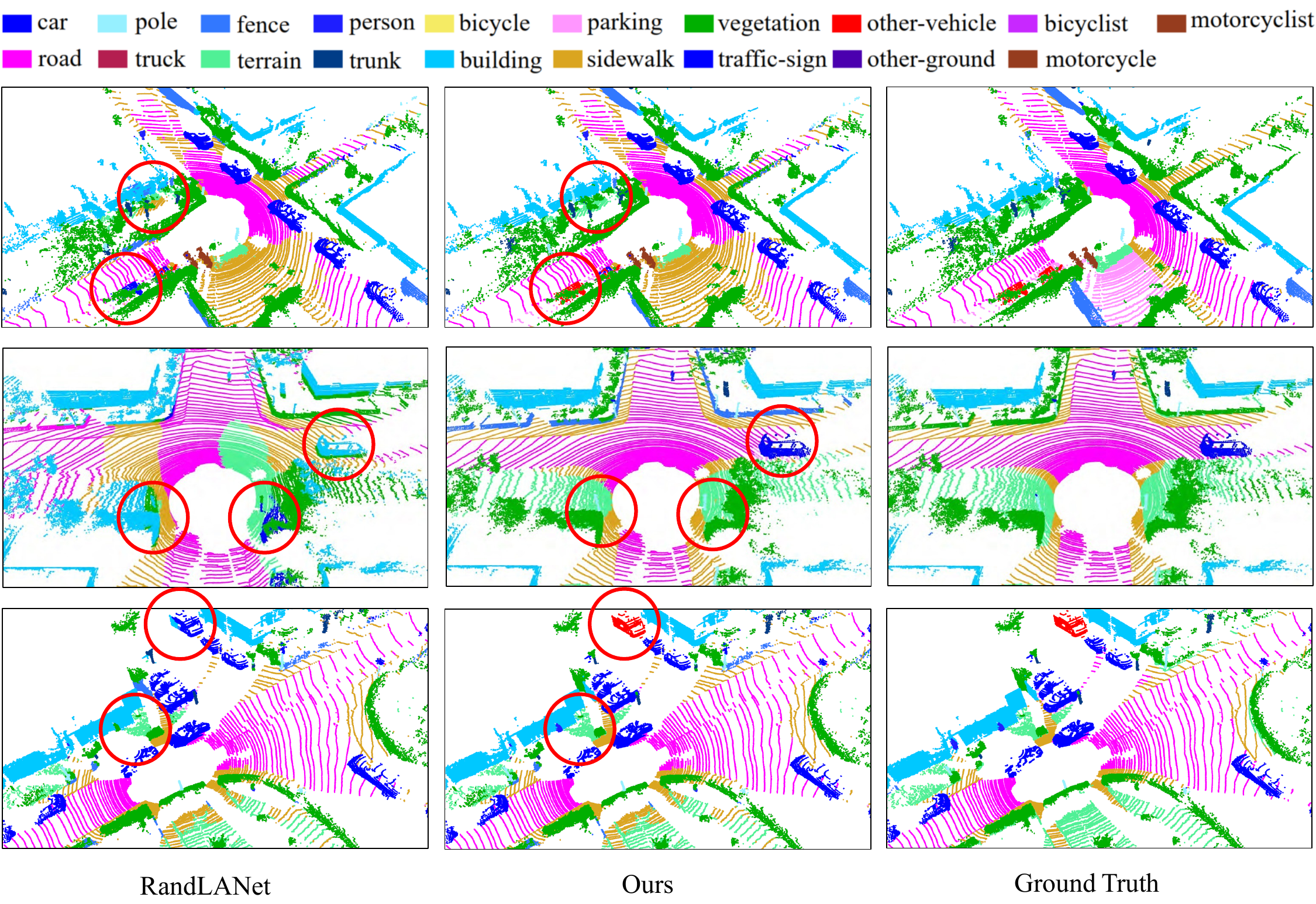}
    \caption{Semantic segmentation results of RandLANet~\cite{hu2020randla}, ours and ground truth on SemanticKITTI dataset~\cite{behley2019semantickitti}.}
    \label{fig:visual_semantickitti}
\end{figure*}

\subsection{Semantic Segmentation on SemanticKITTI}
We provide qualitative results of semantic segmentation on SemanticKITTI dataset in Figure~\ref{fig:visual_semantickitti}.

\subsection{Part Segmentation on ShapeNet dataset}
\begin{figure*}[t]
\centering
\def\sc{0.175}

\includegraphics[width=\sc\linewidth]{figures/visualization/02691156_10db820f0e20396a492c7ca609cb0182_point.png}
\includegraphics[width=\sc\linewidth]{figures/visualization/mesh_airplane.png}
\includegraphics[width=\sc\linewidth]{figures/visualization/02691156_10db820f0e20396a492c7ca609cb0182_combine.png}
\includegraphics[width=\sc\linewidth]{figures/visualization/02691156_10db820f0e20396a492c7ca609cb0182_diff_combine.png}
\includegraphics[width=\sc\linewidth]{figures/visualization/02691156_10db820f0e20396a492c7ca609cb0182_gt.png}

\includegraphics[width=\sc\linewidth]{figures/visualization/02773838_2970e5486815e72ccd99ccc7ff441abf_point.png}
\includegraphics[width=\sc\linewidth]{figures/visualization/mesh_bag.png}
\includegraphics[width=\sc\linewidth]{figures/visualization/02773838_2970e5486815e72ccd99ccc7ff441abf_combine.png}
\includegraphics[width=\sc\linewidth]{figures/visualization/02773838_2970e5486815e72ccd99ccc7ff441abf_diff_combine.png}
\includegraphics[width=\sc\linewidth]{figures/visualization/02773838_2970e5486815e72ccd99ccc7ff441abf_gt.png}

\includegraphics[width=\sc\linewidth]{figures/visualization/02954340_357c2a333ffefc3e90f80ab08ae6ce2_point.png}
\includegraphics[width=\sc\linewidth]{figures/visualization/mesh_hat.png}
\includegraphics[width=\sc\linewidth]{figures/visualization/02954340_357c2a333ffefc3e90f80ab08ae6ce2_combine.png}
\includegraphics[width=\sc\linewidth]{figures/visualization/02954340_357c2a333ffefc3e90f80ab08ae6ce2_diff_combine.png}
\includegraphics[width=\sc\linewidth]{figures/visualization/02954340_357c2a333ffefc3e90f80ab08ae6ce2_gt.png}

\includegraphics[width=\sc\linewidth]{figures/visualization/02958343_f8857237df1717e3aa562f24645e326_point.png}
\includegraphics[width=\sc\linewidth]{figures/visualization/mesh_car.png}
\includegraphics[width=\sc\linewidth]{figures/visualization/02958343_f8857237df1717e3aa562f24645e326_combine.png}
\includegraphics[width=\sc\linewidth]{figures/visualization/02958343_f8857237df1717e3aa562f24645e326_diff_combine.png}
\includegraphics[width=\sc\linewidth]{figures/visualization/02958343_f8857237df1717e3aa562f24645e326_gt.png}

\vspace{0.2in}
\includegraphics[width=\sc\linewidth]{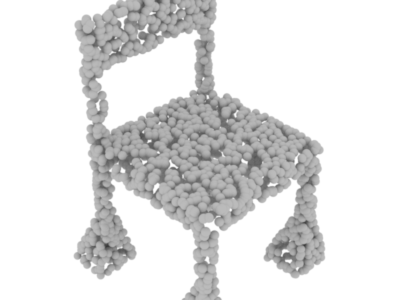}
\includegraphics[width=\sc\linewidth]{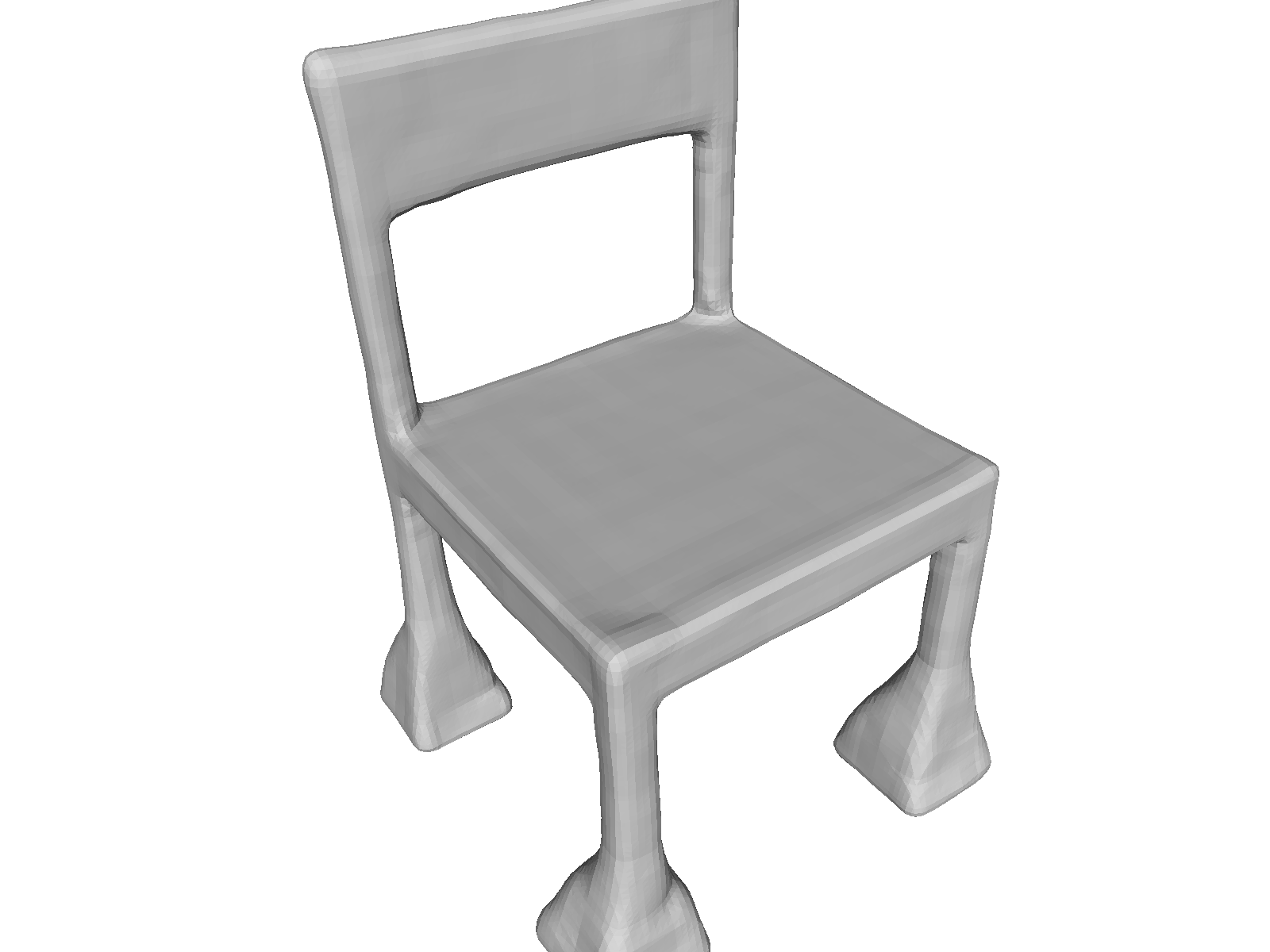}
\includegraphics[width=\sc\linewidth]{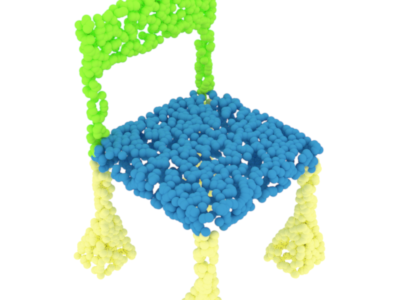}
\includegraphics[width=\sc\linewidth]{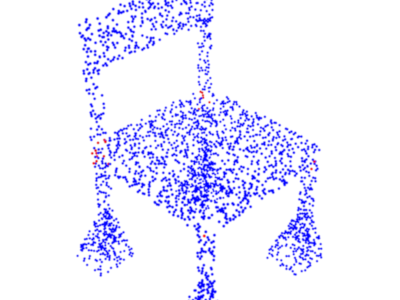}
\includegraphics[width=\sc\linewidth]{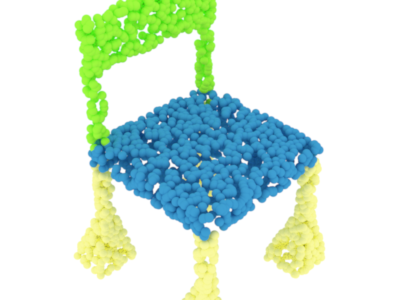}

\includegraphics[width=\sc\linewidth]{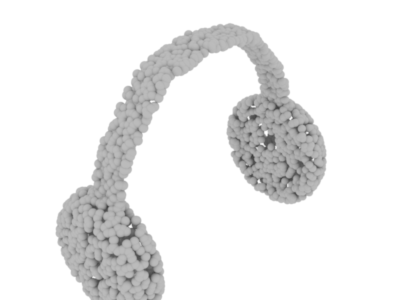}
\includegraphics[width=\sc\linewidth]{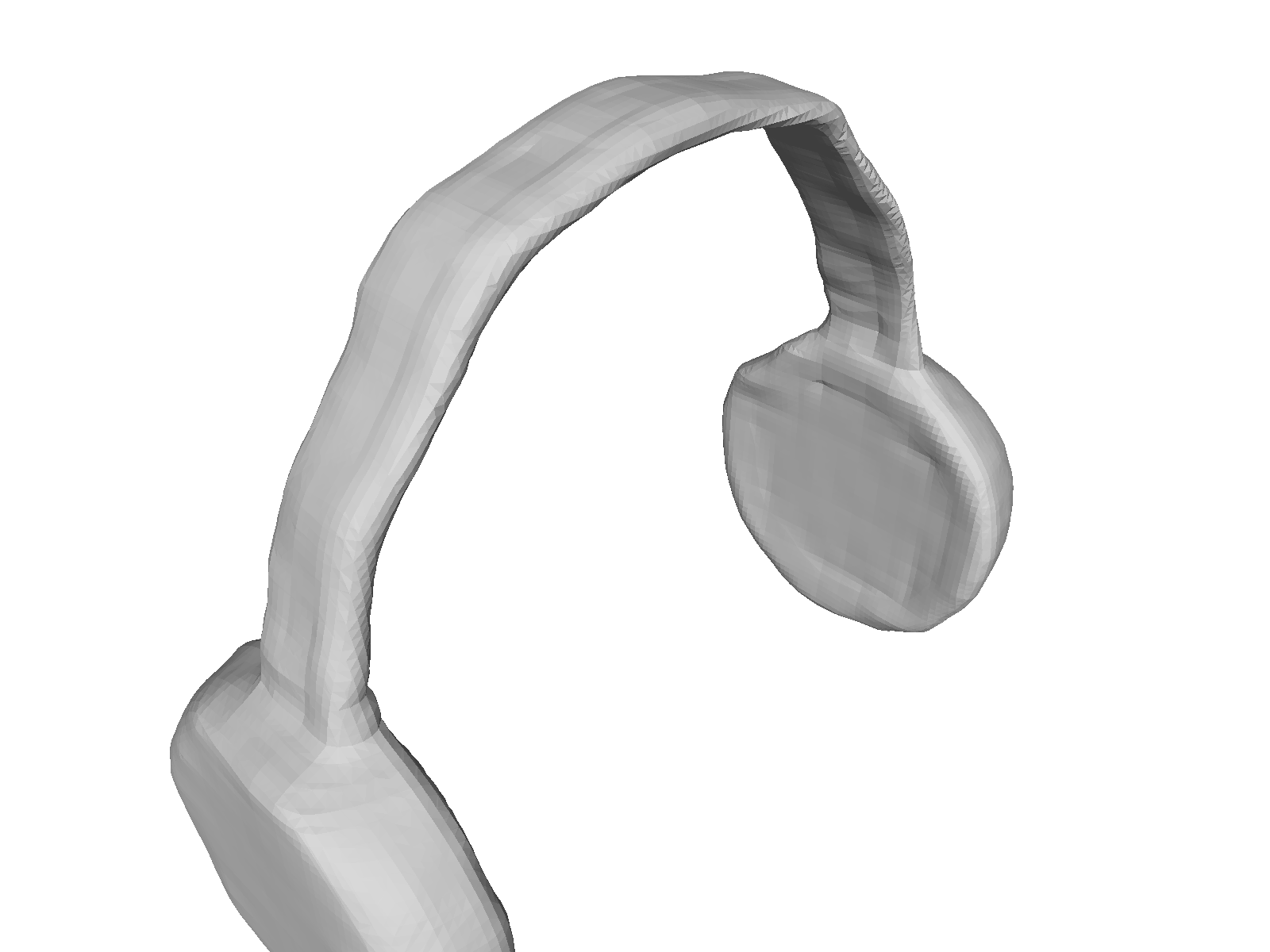}
\includegraphics[width=\sc\linewidth]{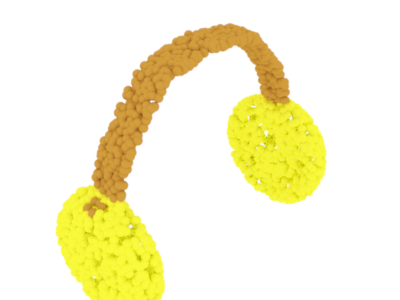}
\includegraphics[width=\sc\linewidth]{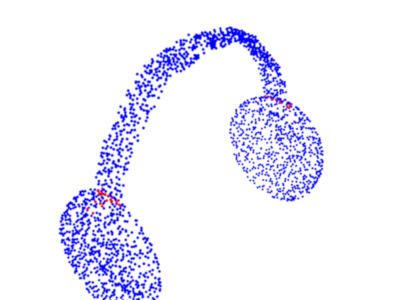}
\includegraphics[width=\sc\linewidth]{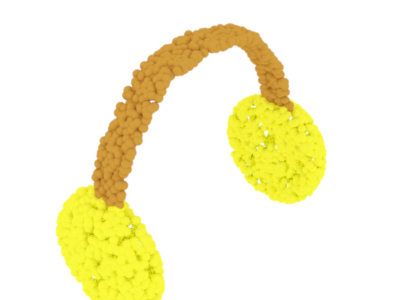}

\vspace{0.1in}
\includegraphics[width=\sc\linewidth]{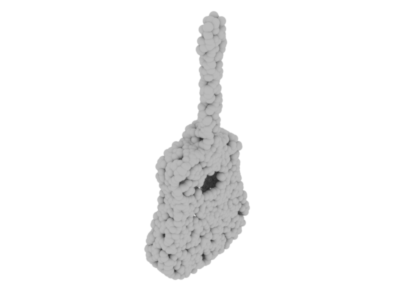}
\includegraphics[width=\sc\linewidth]{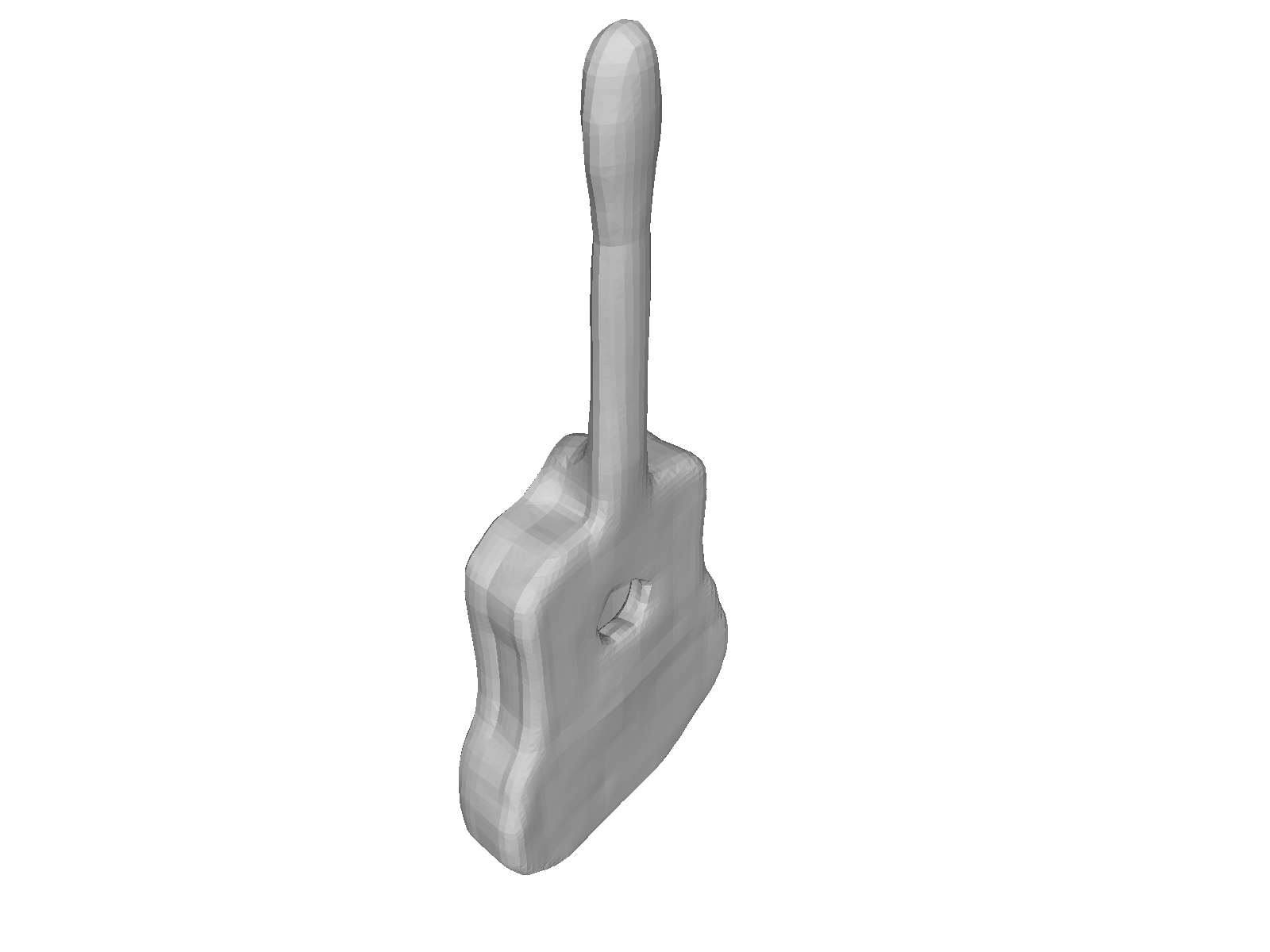}
\includegraphics[width=\sc\linewidth]{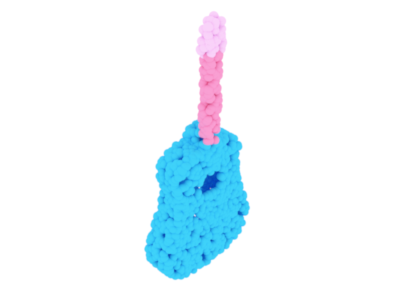}
\includegraphics[width=\sc\linewidth]{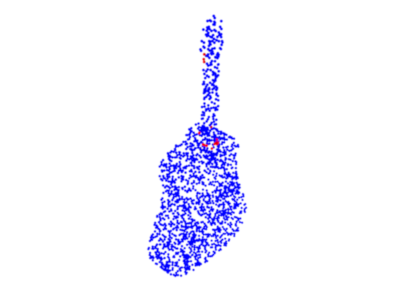}
\includegraphics[width=\sc\linewidth]{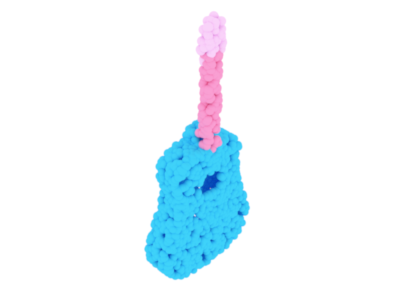}

\vspace{0.2in}
\includegraphics[width=\sc\linewidth]{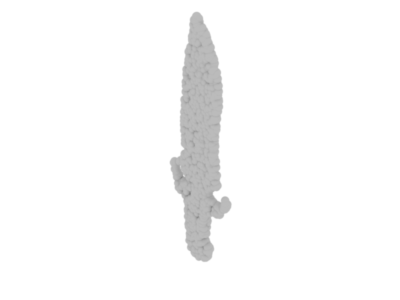}
\includegraphics[width=\sc\linewidth]{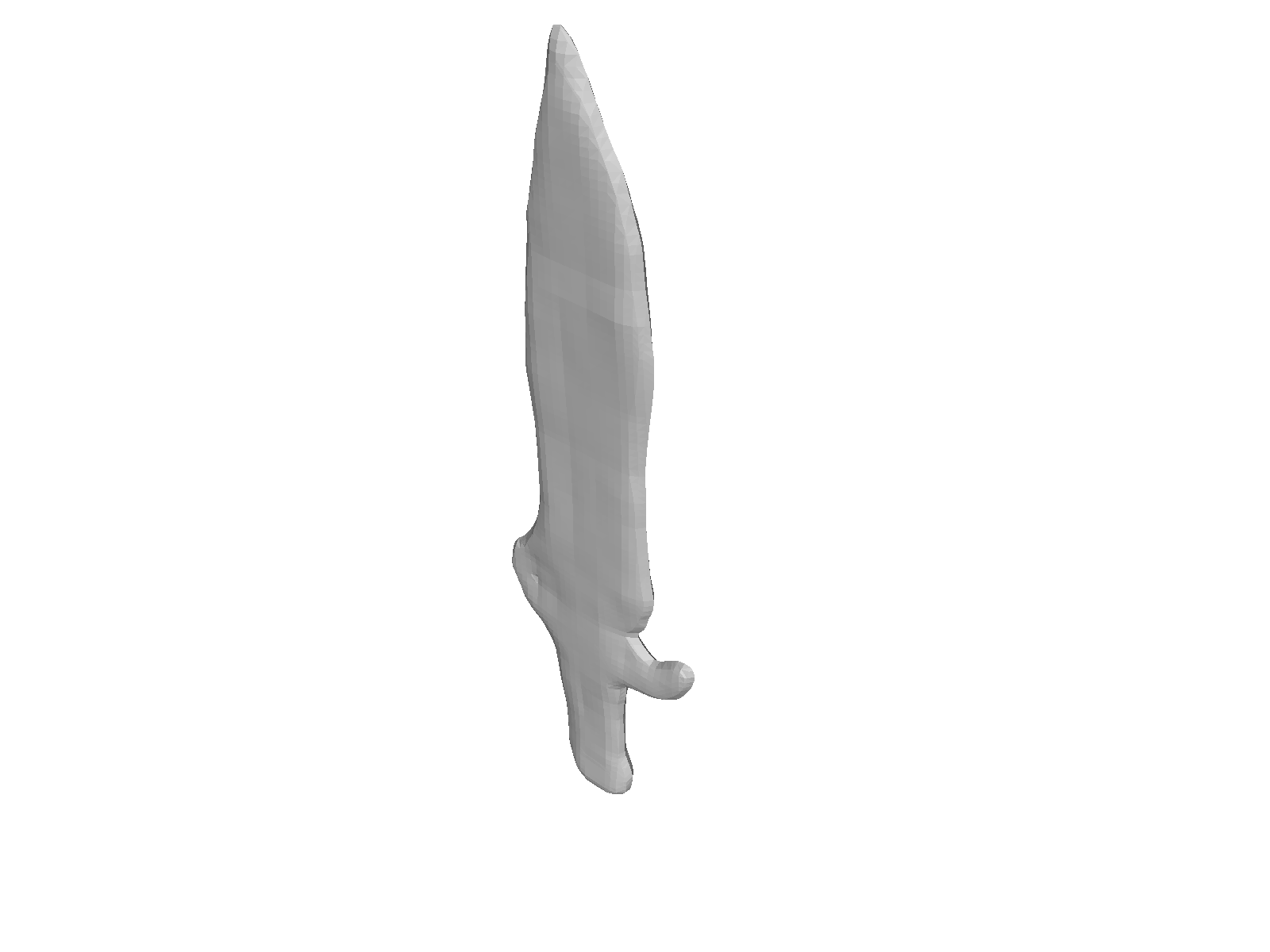}
\includegraphics[width=\sc\linewidth]{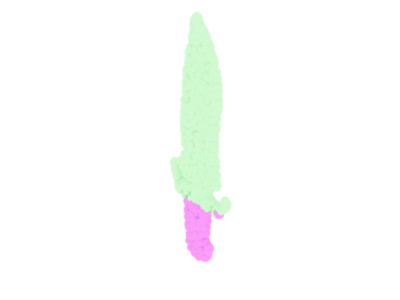}
\includegraphics[width=\sc\linewidth]{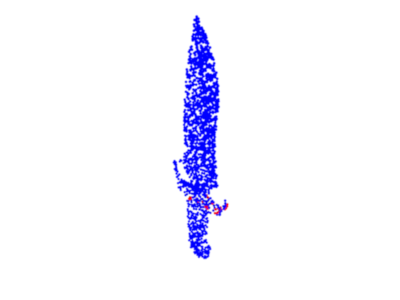}
\includegraphics[width=\sc\linewidth]{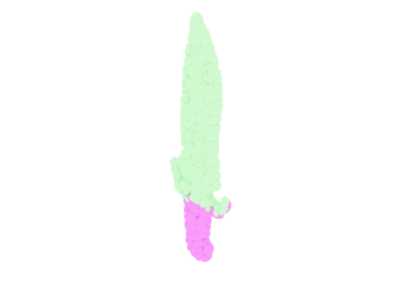}

\centerline{\hspace{0.3in} (a) Input \hspace{0.6in} (b) Reconstruction \hspace{0.5in} (c) Our result \hspace{0.5in} (d) Difference map \hspace{0.3in} (e) Ground truth}
\vspace{0.1in}
\caption{Visualization of part segmentation results. As can be seen in the difference map, where blue and red points indicate correct and wrong labels, respectively, our test-time augmentation mainly deals with the labels along the boundaries, improving their accuracies through aggregating predictions from augmented point clouds. Best viewed with zoom.}
\label{fig:part_seg_point_1}
\end{figure*}

\begin{figure*}[t]
\centering
\def\sc{0.175}

\includegraphics[width=\sc\linewidth]{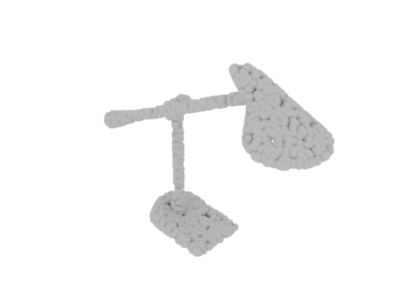}
\includegraphics[width=\sc\linewidth]{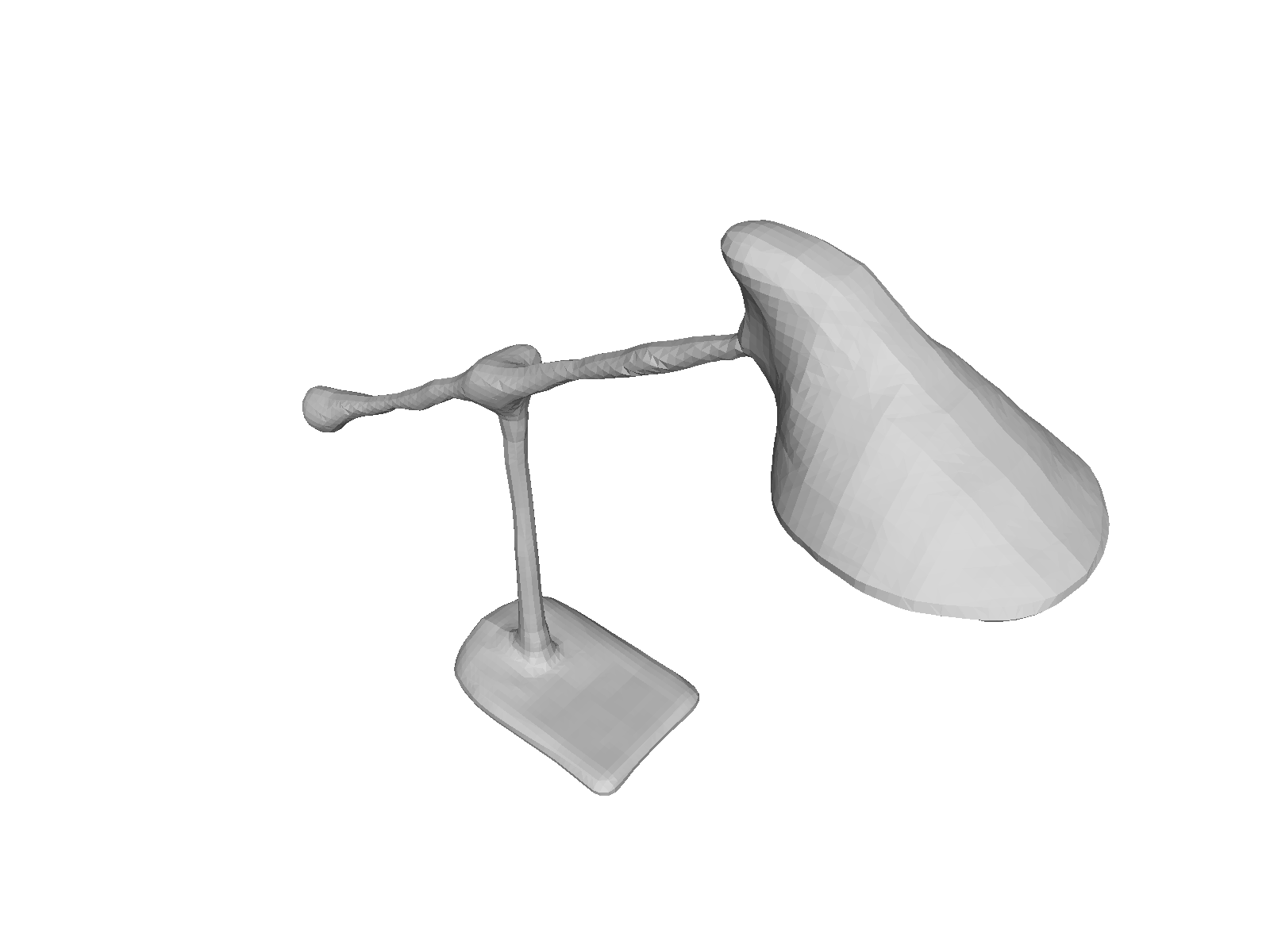}
\includegraphics[width=\sc\linewidth]{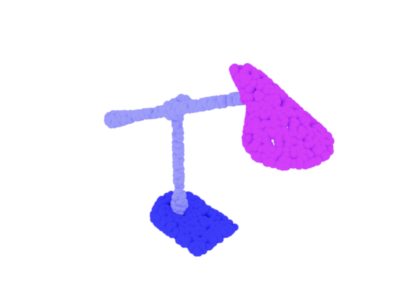}
\includegraphics[width=\sc\linewidth]{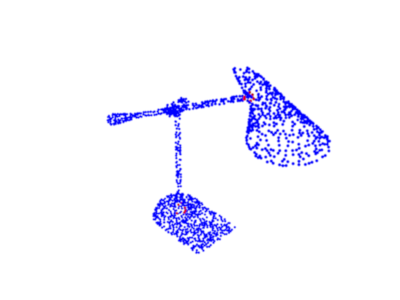}
\includegraphics[width=\sc\linewidth]{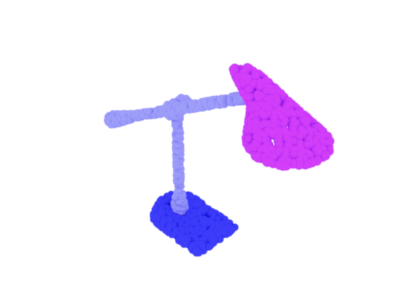}

\vspace{0.1in}
\includegraphics[width=\sc\linewidth]{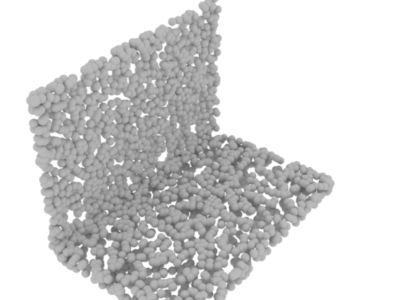}
\includegraphics[width=\sc\linewidth]{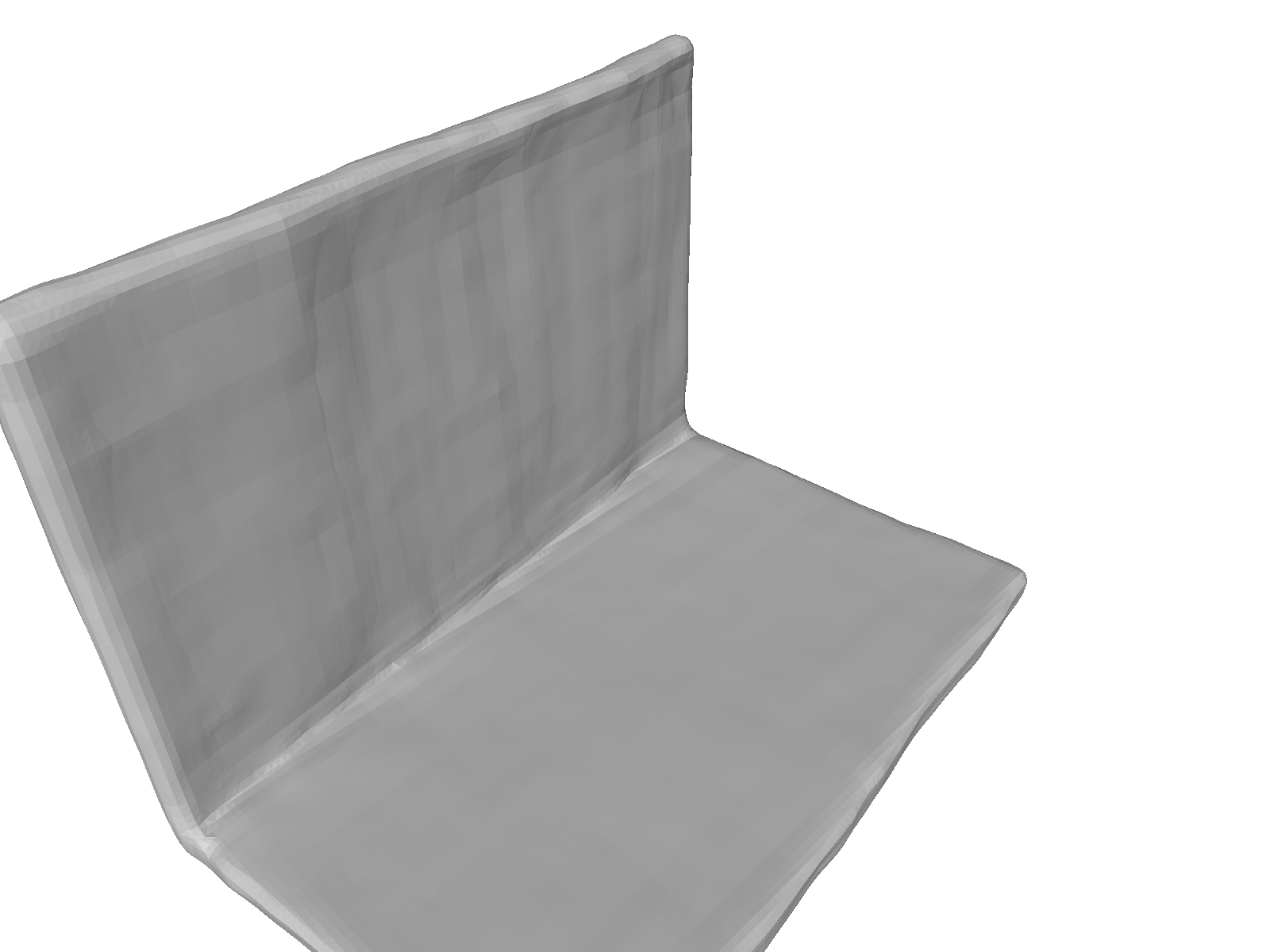}
\includegraphics[width=\sc\linewidth]{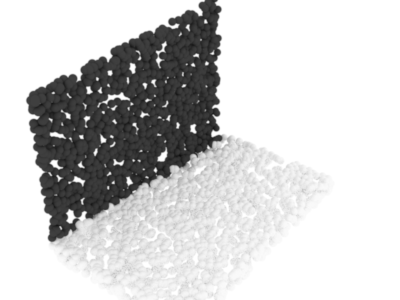}
\includegraphics[width=\sc\linewidth]{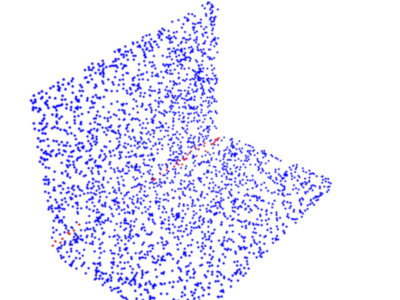}
\includegraphics[width=\sc\linewidth]{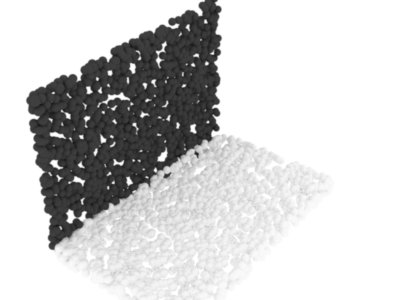}

\includegraphics[width=\sc\linewidth]{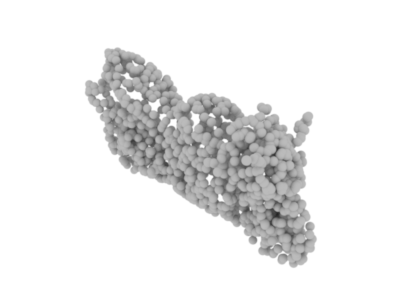}
\includegraphics[width=\sc\linewidth]{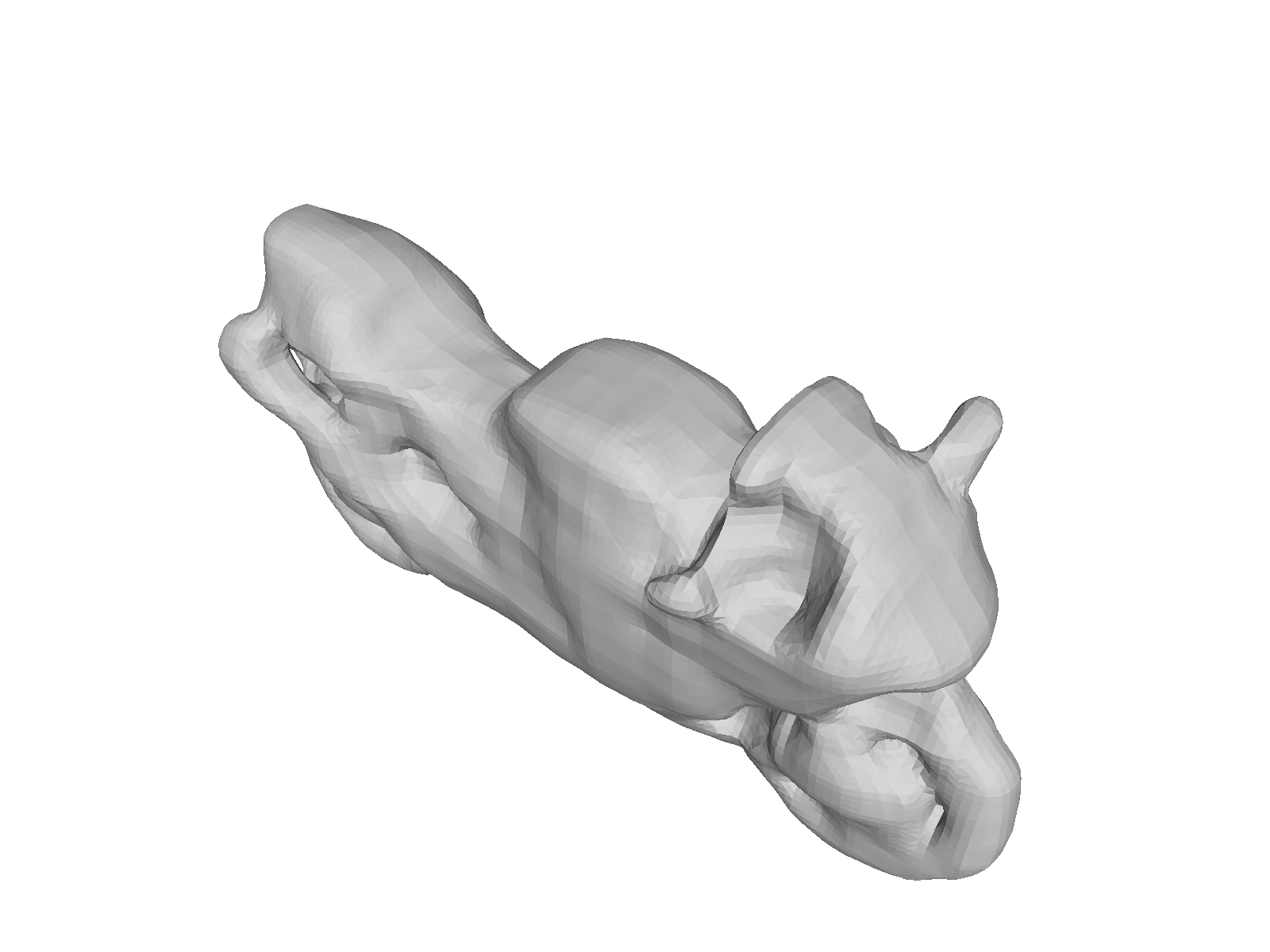}
\includegraphics[width=\sc\linewidth]{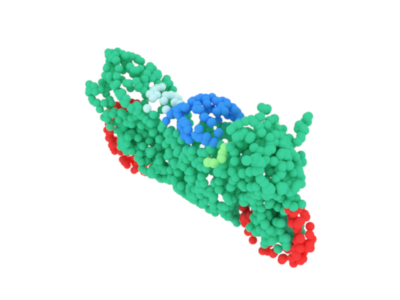}
\includegraphics[width=\sc\linewidth]{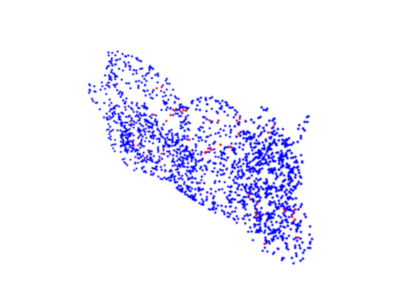}
\includegraphics[width=\sc\linewidth]{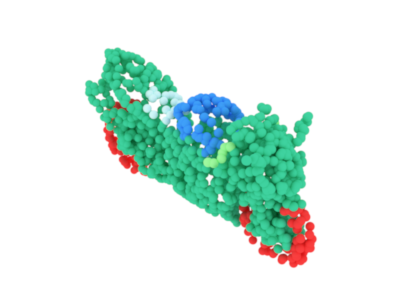}

\vspace{0.2in}
\includegraphics[width=\sc\linewidth]{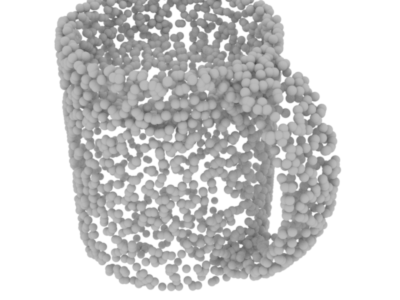}
\includegraphics[width=\sc\linewidth]{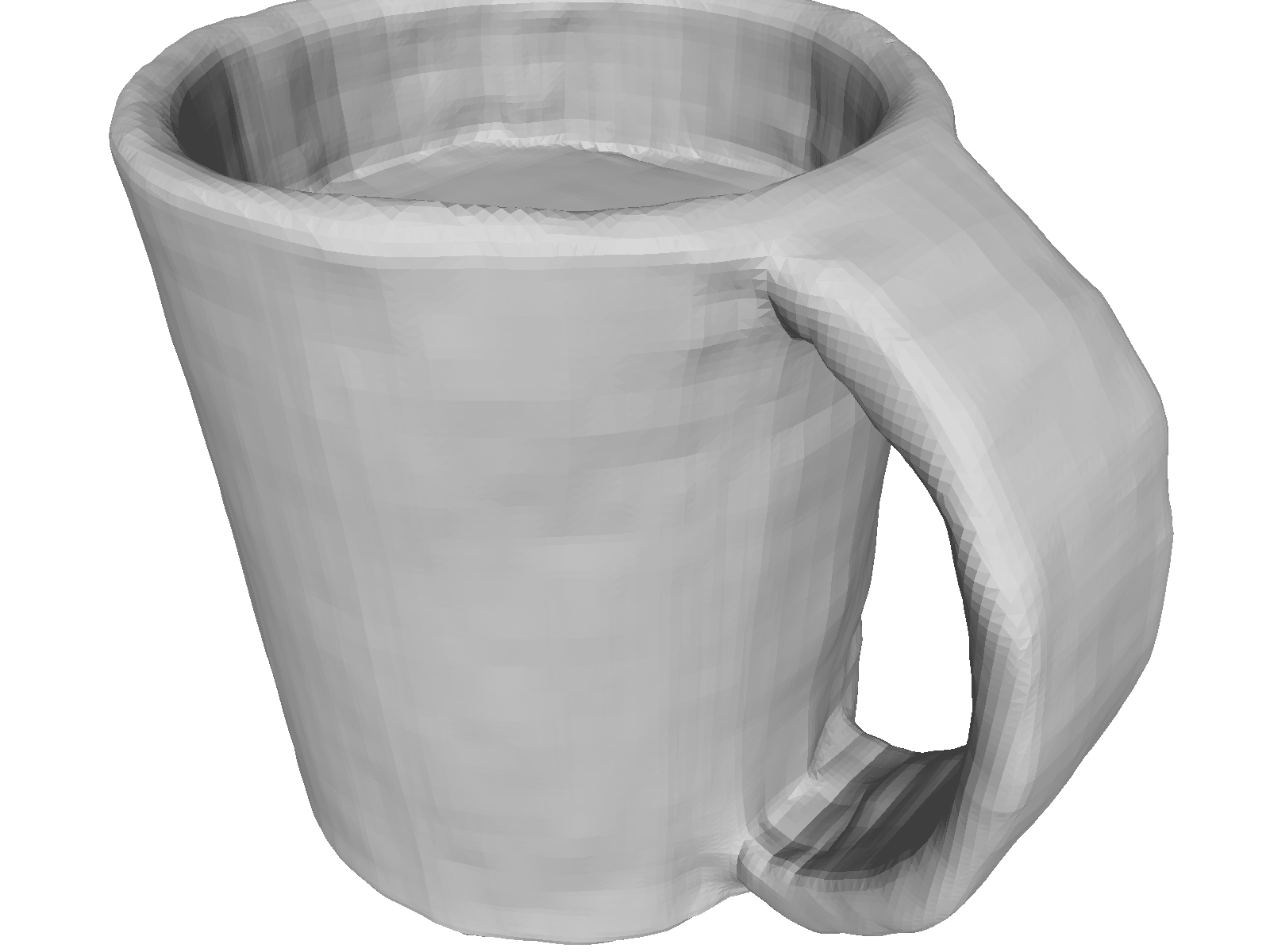}
\includegraphics[width=\sc\linewidth]{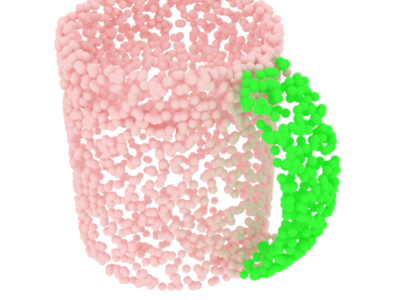}
\includegraphics[width=\sc\linewidth]{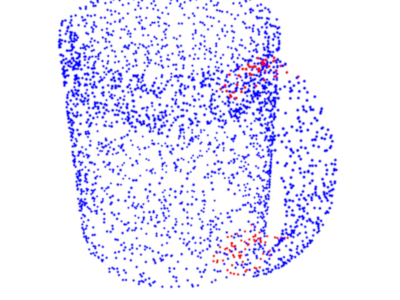}
\includegraphics[width=\sc\linewidth]{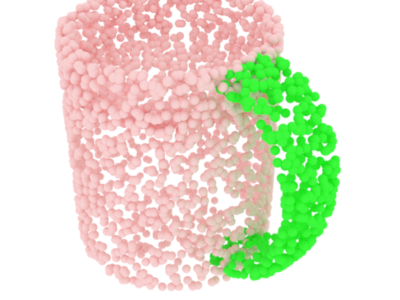}

\includegraphics[width=\sc\linewidth]{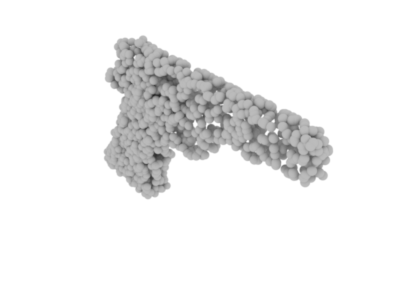}
\includegraphics[width=\sc\linewidth]{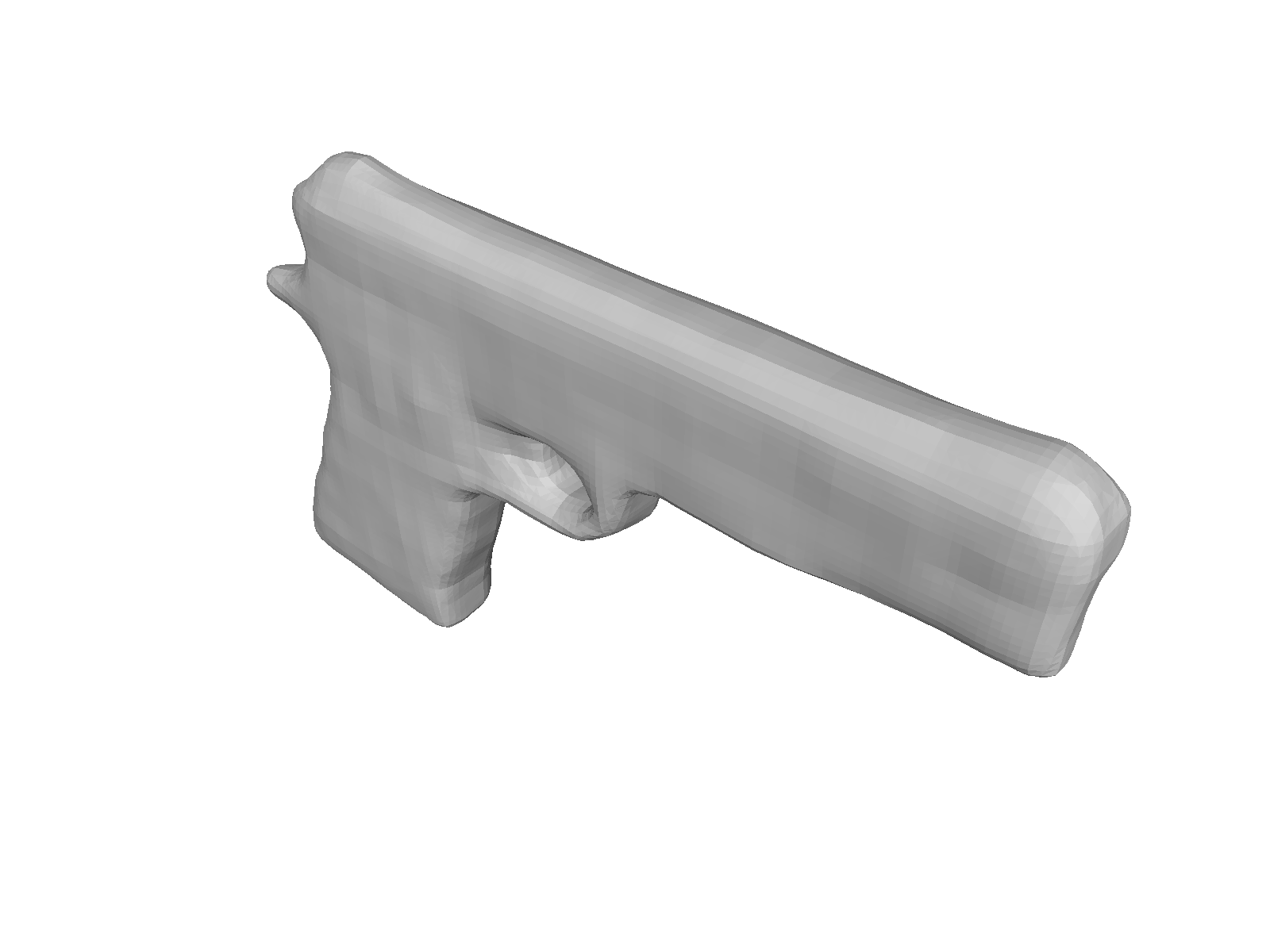}
\includegraphics[width=\sc\linewidth]{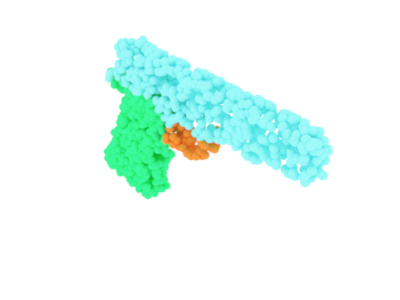}
\includegraphics[width=\sc\linewidth]{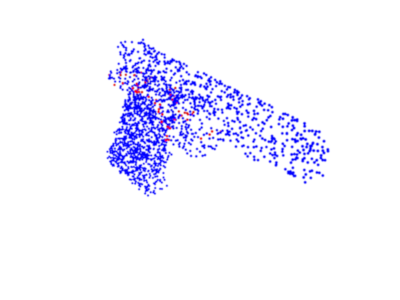}
\includegraphics[width=\sc\linewidth]{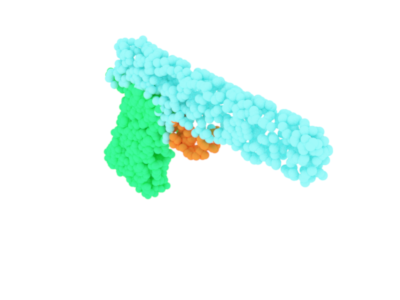}

\includegraphics[width=\sc\linewidth]{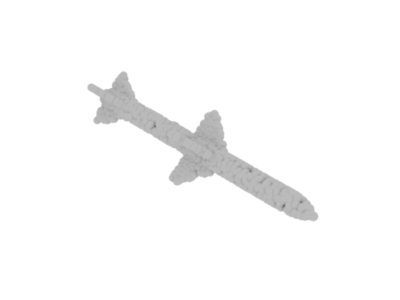}
\includegraphics[width=\sc\linewidth]{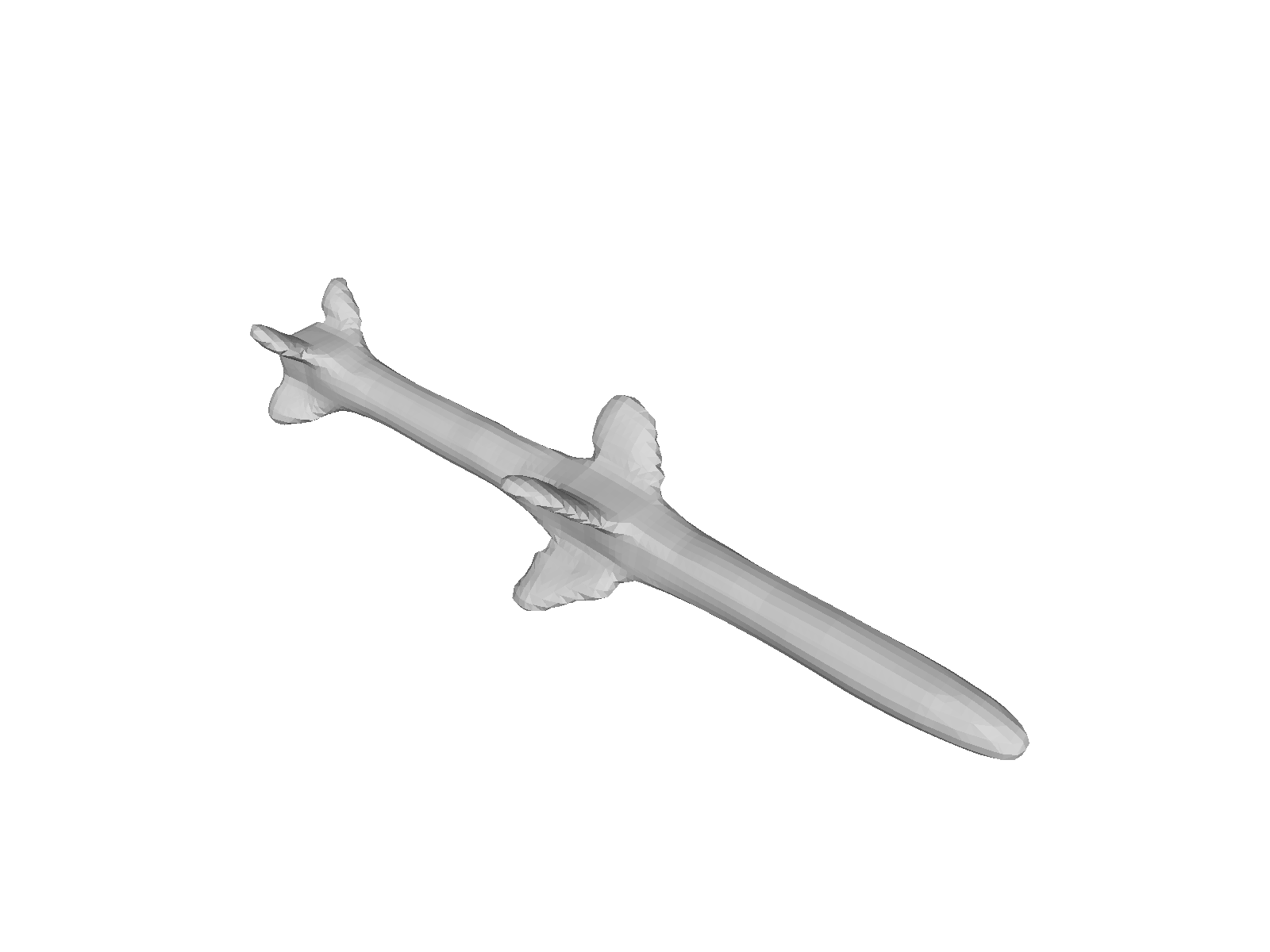}
\includegraphics[width=\sc\linewidth]{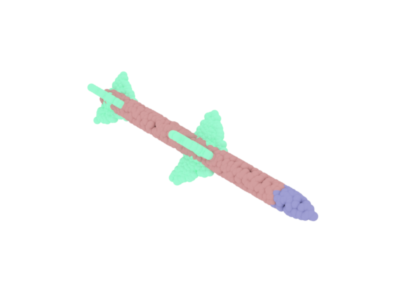}
\includegraphics[width=\sc\linewidth]{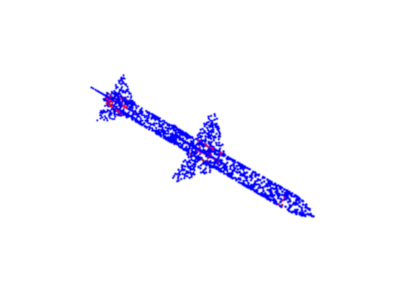}
\includegraphics[width=\sc\linewidth]{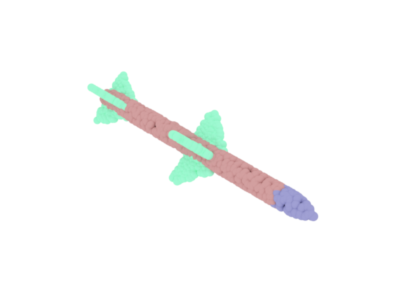}

\includegraphics[width=\sc\linewidth]{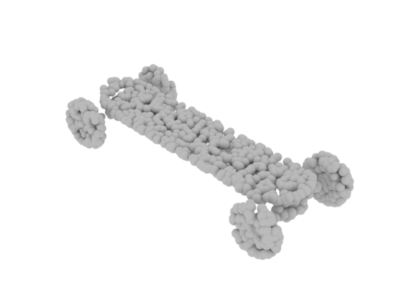}
\includegraphics[width=\sc\linewidth]{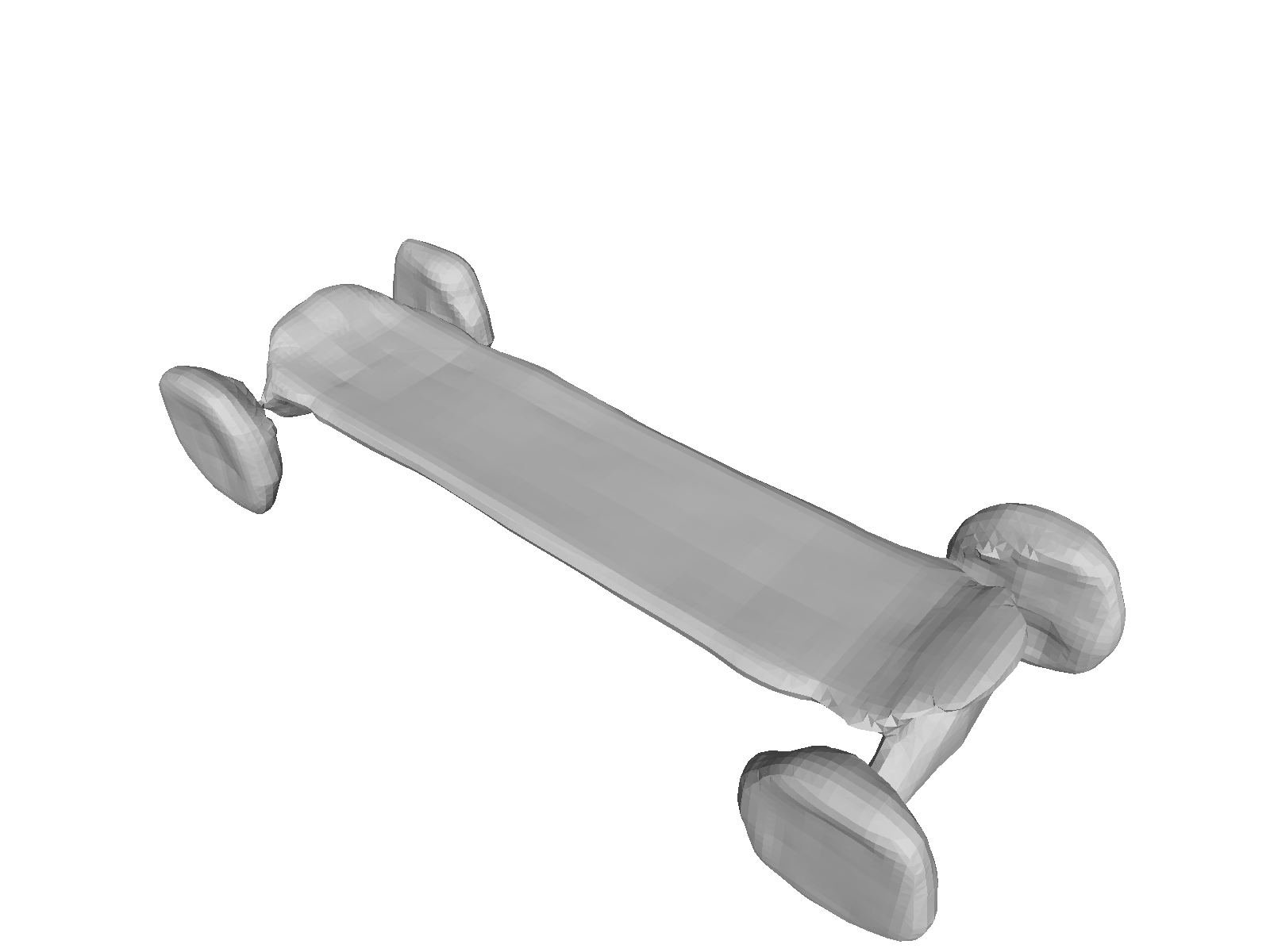}
\includegraphics[width=\sc\linewidth]{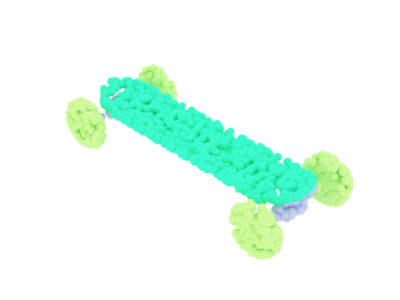}
\includegraphics[width=\sc\linewidth]{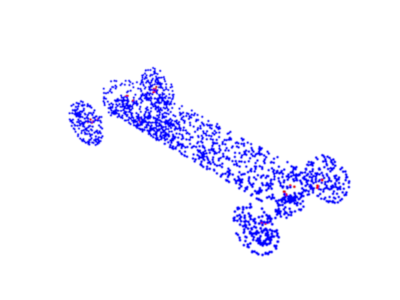}
\includegraphics[width=\sc\linewidth]{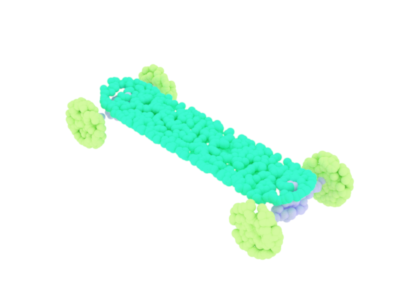}

\vspace{0.2in}
\includegraphics[width=\sc\linewidth]{figures/visualization/04379243_855e0dd7d801faf61886830ef08153db_point.png}
\includegraphics[width=\sc\linewidth]{figures/visualization/mesh_table.png}
\includegraphics[width=\sc\linewidth]{figures/visualization/04379243_855e0dd7d801faf61886830ef08153db_combine.png}
\includegraphics[width=\sc\linewidth]{figures/visualization/04379243_855e0dd7d801faf61886830ef08153db_diff_combine.png}
\includegraphics[width=\sc\linewidth]{figures/visualization/04379243_855e0dd7d801faf61886830ef08153db_gt.png}

\centerline{\hspace{0.3in} (a) Input \hspace{0.6in} (b) Reconstruction \hspace{0.5in} (c) Our result \hspace{0.5in} (d) Difference map \hspace{0.3in} (e) Ground truth}
\vspace{0.1in}
\caption{Visualization of part segmentation results. As can be seen in the difference map, where blue and red points indicate correct and wrong labels, respectively, our test-time augmentation mainly deals with the labels along the boundaries, improving their accuracies through aggregating predictions from augmented point clouds. Best viewed with zoom.}
\label{fig:part_seg_point_2}
\end{figure*}

We provide more qualitative results of part segmentation on ShapeNet dataset in Figure~\ref{fig:part_seg_point_1} and Figure~\ref{fig:part_seg_point_2}.

\end{document}